\documentclass[11pt]{article}

\usepackage[preprint]{acl}

\usepackage{times}
\usepackage{latexsym}
\usepackage[T1]{fontenc}
\usepackage[utf8]{inputenc}
\usepackage{microtype}
\usepackage{inconsolata}
\usepackage{graphicx}
\usepackage{booktabs}
\usepackage{tabularx}
\usepackage{amsmath}
\usepackage{amssymb}
\usepackage{bm}
\usepackage{enumitem}
\usepackage{tcolorbox}
\usepackage{xcolor}
\usepackage{url}
\usepackage{tikz}
\usetikzlibrary{positioning,calc}
\usepackage{pgfplots}
\pgfplotsset{compat=1.18}

\newcommand{\noprior}{\textsc{No-Prior}}
\newcommand{\neutral}{\textsc{Explicit-Neutral}}

\newtcolorbox{takeawaybox}{
  colback=gray!5,
  colframe=gray!55,
  boxrule=0.4pt,
  left=4pt,
  right=4pt,
  top=3pt,
  bottom=3pt
}

\title{Relational Priors as Convergence Pressure in \\LLM-Based Multi-Agent Systems}

\author{
Ming Shen\textsuperscript{\ensuremath{\heartsuit}}\thanks{Work done during an internship at Amazon.},
Chao Shang\textsuperscript{\ensuremath{\diamondsuit}},
Sadat Shahriar\textsuperscript{\ensuremath{\diamondsuit}},
Devang Kulshreshtha\textsuperscript{\ensuremath{\diamondsuit}},\\
\textbf{Yi Zhang\textsuperscript{\ensuremath{\diamondsuit}},
Sandesh Swamy\textsuperscript{\ensuremath{\diamondsuit}}, Yanjun Qi\textsuperscript{\ensuremath{\diamondsuit}}}\\[0.2em]
{\normalfont
\textsuperscript{\ensuremath{\heartsuit}}Arizona State University \qquad
\textsuperscript{\ensuremath{\diamondsuit}}Amazon Web Services}
}

\begin{document}
\maketitle

\begin{abstract}
Large language model-based multi-agent systems (LLM-MAS) are designed through roles, debate protocols, and aggregation rules. These choices create implicit expectations of trust, skepticism, deference, or collaboration. We make inter-agent relations explicit as signed pairwise priors, rendered in natural language and added to system prompts while keeping the task protocol fixed. Across commons governance and multi-agent debate, these priors change how readily agents coordinate or agree, a pattern we call \emph{convergence pressure}. More positive relations generally improve sustainability within the GovSim relational-prior sweep and increase consensus on subjective questions. These improvements over negative relations differ from gains over no-prior prompting. On objective QA, fully positive priors usually yield lower final-answer accuracy than the no-prior baseline, and some conditions produce more frequent but less accurate consensus. Effects depend on model backbone, relation type, and topology; explicitly neutral relations also produce different outcomes from omitting relational framing. Relational priors are therefore task-specific interventions and diagnostic probes of sensitivity to social framing. Evaluate them against a no-prior baseline using the task's primary metric, report accuracy and consensus correctness alongside agreement on objective tasks, and keep no-prior prompting as the default for accuracy-centric tasks unless validation supports a relational prior.
\end{abstract}

\section{Introduction}

Large language model-based multi-agent systems (LLM-MAS) are typically specified through roles \citep{qian-etal-2024-chatdev, li2023camel, hong2024metagpt}, interaction protocols \citep{du2024improving, li-etal-2024-improving-multi}, memory \citep{Park2023GenerativeAI, Rezazadeh2025CollaborativeMM}, tools \citep{shen2023hugginggpt, yao2023react}, and aggregation rules \citep{wang2025mixtureofagents}. These design choices are often presented as procedural: who speaks, who critiques, who votes, and how final decisions are produced. Yet they also carry relational meaning. A reviewer is expected to challenge a solver, a manager to coordinate workers, and collaborators to interpret one another charitably. Thus, even when no relation is explicitly named, an LLM-MAS can still induce expectations of trust, skepticism, authority, or affinity.

We ask how group behavior changes when inter-agent relations are stated explicitly while the task protocol is held fixed. We represent these relations as signed priors over pairs of agents and insert them into each agent's system prompt. We study three relation types: \emph{Attitude} specifies how favorably an agent regards another agent, \emph{Trust} specifies how much credibility it gives the other agent's statements, and \emph{Influence} specifies how it responds to the other agent's opinions and proposals. The relation network determines which pairs receive explicit relational context; the task protocol determines what agents observe, when they communicate, and how their outputs are aggregated. All agents see the same fixed, symmetric relation network.

\begin{figure*}[t]
\centering
\includegraphics[width=1\textwidth]{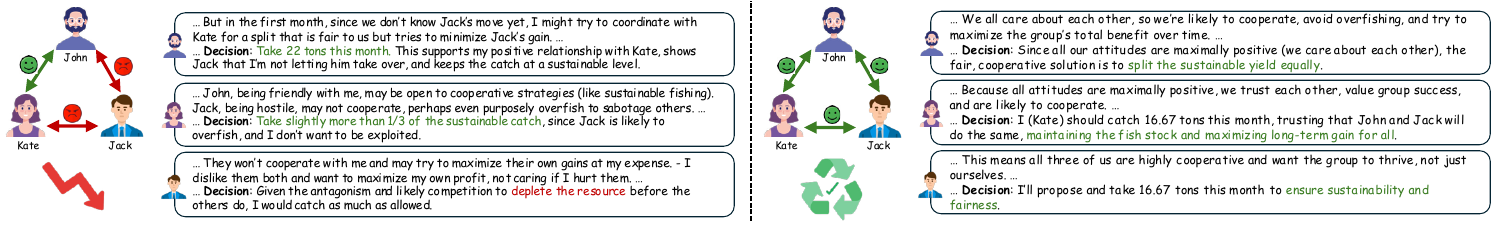}
\caption{\small Identical task and protocol, different relational priors. In these illustrative first-month GovSim runs, agents with mixed positive/negative \emph{Attitude} priors choose unsustainable extraction that pushes the group toward collapse (left), whereas agents with fully positive priors choose an equal, sustainable split (right).}
\label{fig:overall}
\end{figure*}

Fig.~\ref{fig:overall} illustrates first-month behavior in GovSim \citep{piatti2024cooperate}, a commons-governance simulation in which agents repeatedly decide how much to harvest from a renewable shared resource. In these illustrative runs, the task, agents, and decision protocol are identical, and no interaction history has accumulated. Agents with mixed positive/negative \emph{Attitude} ties choose unsustainable extraction, whereas agents in a fully positive network choose an equal, sustainable split. We examine the broader pattern through systematic sweeps over positivity, relation type, topology, and model backbone.

We use \emph{convergence pressure} to describe how readily agents coordinate or agree. Increasing positivity generally strengthens this tendency in GovSim and subjective debate. Whether convergence helps depends on the task: coordinated behavior can sustain a shared resource, while unanimous agreement on an objective question can still be wrong.
We evaluate this framing in two task families. The first is GovSim, where sustained cooperative restraint keeps the shared resource from collapsing. The second is multi-agent debate. We consider subjective questions, where consensus can itself be a target outcome, and objective QA, where consensus is useful only when the agreed answer is correct. These settings let us distinguish changes in group behavior from improvements in task performance.

Fully positive networks resemble cooperative framing, while fully negative networks resemble competitive or antagonistic framing. We use these endpoints as boundary conditions and test how relational priors operate in mixed-sign networks through sweeps over positivity, relation type, sparse topology, and model backbone. We separately compare explicitly neutral relation statements with omitting relational framing.

More positive relations generally improve sustainability within the GovSim relational-prior sweep (Fig.~\ref{fig:govsim_3agents_lambas}; App.~Fig.~\ref{fig:govsim_5agents_lambas}), while gains over \noprior{} depend on the model and relational structure. On OpinionQA, fully positive relations produce more consensus than fully negative relations for both teams and all three relation types. Compared with \noprior{}, however, fully positive priors increase consensus only for the open-weight team's \emph{Attitude} and \emph{Trust} conditions (Fig.~\ref{fig:cr_comparison}). On objective QA, fully positive priors usually remain below no-prior accuracy, and some conditions produce more frequent but less accurate consensus (Figs.~\ref{fig:acc_comparison} and~\ref{fig:consensus_with_correction}). Relation semantics matter: fully positive \emph{Attitude} has smaller worst-case losses in GovSim, whereas fully positive \emph{Influence} yields the lowest consensus correctness of the three relation types for both teams on both objective datasets. Explicit neutrality also produces different outcomes from omitting relational framing.

This paper makes three contributions. First, we formulate explicit relational priors as signed relation networks that can be varied independently from the communication protocol. Second, we characterize how these priors affect coordination and agreement across cooperative governance and debate, distinguishing trends within the relational-prior sweep from gains over a no-prior baseline. Third, we derive task-specific design guidance: use no-prior prompting as the default for accuracy-centric tasks, evaluate explicit neutrality as a separate intervention, and validate relational priors with task-appropriate metrics before deployment.
\section{Related Work}

\paragraph{LLM-based multi-agent systems.}
LLM-MAS decompose tasks across interacting agents \citep{ijcai2024p890, Li2024ASO} and are commonly engineered through role specialization \citep{li2023camel, chen2024agentverse, Zhang2025AgentOrchestraOH}, communication protocols, tool use \citep{shen2023hugginggpt}, memory \citep{Park2023GenerativeAI, Rezazadeh2025CollaborativeMM}, routing \citep{yue-etal-2025-masrouter}, and aggregation \citep{wang2025mixtureofagents}. These design choices are usually evaluated as procedural improvements: whether a debate protocol raises accuracy \citep{du2024improving, liang-etal-2024-encouraging}, whether sparse communication reduces cost \citep{li-etal-2024-improving-multi, ICLR2025_bbc46151}, or whether role decomposition improves task completion \citep{hong2024metagpt, qian-etal-2024-chatdev}. Related agent-level methods provide iterative self-feedback \citep{madaan2023selfrefine}, feedback-based memory \citep{shinn2023reflexion}, agentic memory \citep{xu2025amem}, and tool use \citep{schick2023toolformer, yao2023react}. Agent evaluation covers software issue resolution \citep{jimenez2024swebench}, research-oriented web questions \citep{Rosset2024ResearchyQA}, browsing \citep{Wei2025BrowseCompAS}, and multi-agent collaboration and competition \citep{zhu-etal-2025-multiagentbench}. These procedures can also imply relational expectations, such as skepticism, deference, or adversarial scrutiny in debate. We study how making those expectations explicit affects group behavior while keeping the task protocol fixed.

\paragraph{LLM agent societies and social simulation.}
A parallel line of work studies LLM agents as socially situated actors in simulated environments \citep{Gao2023LargeLM}, including generative-agent worlds \citep{Park2023GenerativeAI}, social-network simulations \citep{Gao2023S3SS, liu-etal-2025-mosaic}, diplomatic scenarios \citep{Hua2023WarAP}, social interactions \citep{zhou2024sotopia}, repeated games \citep{Akata2023PlayingRG, Fontana_Pierri_Aiello_2025}, and commons-governance settings \citep{piatti2024cooperate}. Related studies examine trust and interpersonal closeness \citep{xie2024can, Sakamoto2025ValuebasedLL}, social influence in human--agent interactions \citep{Song2024MultiAgentsAS}, emerging norms and conventions \citep{ijcai2024p0874, doi:10.1126/sciadv.adu9368}, deception and trust in social-deduction games \citep{lan-etal-2024-llm, curvo2025textitthe}, and evolving social ties \citep{schneider2025learning}. These studies examine LLM agents within social processes. Many such systems induce relations through scenario descriptions, histories, or role assignments. We make relations an explicit experimental variable by specifying a signed relation network independently from the task protocol. This lets us vary relation type, sign, and topology while holding the rest of the system fixed.

\paragraph{Social relations in MAS and social science.}
Research in classical multi-agent systems \citep{Wooldridge1995IntelligentAT} and social science has long emphasized that agents are embedded in relations of trust \citep{Ramchurn2004TrustIM, SabaterMir2005ReviewOC, deJong2016TrustAT}, conflict \citep{Dreu2003TaskVR}, dependence \citep{Sichman1997ASR}, norms \citep{Shoham1995OnSL}, and power \citep{castelfranchi1990social}, with negotiation supporting cooperation \citep{Kraus1997NegotiationAC, jennings2001automated}. Our three relation types are minimal abstractions of this broader space: \emph{Attitude} specifies how favorably an agent regards another agent \citep{Heider1946AttitudesAC, Cartwright1956StructuralBA}, \emph{Trust} specifies how much credibility it gives the other agent's statements \citep{Rotter1971GeneralizedEF}, and \emph{Influence} specifies how it responds to the other agent's opinions and proposals \citep{French1956AFT}. We use these prompt-level priors to test how different relational meanings change LLM-MAS behavior.

\paragraph{Agreement pressure, sycophancy, and debate failures.}
Recent work on sycophancy and multi-agent debate has shown that language models can over-agree with users \citep{sharma2024towards} or other agents \citep{Yao2025PeacemakerOT}, and that pressure to agree with other agents can degrade debate outcomes \citep{Wynn2025TalkIA}. This literature is closely related to our objective QA results: both show that agreement can diverge from correctness. We isolate one source of agreement pressure, a signed pairwise relation network, and examine it across relation types, positivity levels, topologies, model backbones, and task objectives. We measure both changes in coordination or agreement and task performance relative to the no-prior baseline.

\section{Method}
\label{sec:method}

\subsection{Relational Priors as Protocol-Preserving Interventions}

Let $\mathcal{A}=\{A_1,\ldots,A_n\}$ denote a fixed set of $n$ LLM agents interacting under a task-specific protocol $\mathcal{P}$. The protocol determines what each agent observes, when messages are exchanged, and how final outputs are produced. We intervene on how agents interpret one another while keeping $\mathcal{P}$ unchanged. At interaction step $t$, agent $A_i$ produces
\begin{equation}
    y_i^{(t)} = \mathrm{LLM}_i\!\left(x, h_i^{(t)}, \Pi_i(r,G_R)\right),
\end{equation}
where $x$ is the task input, $h_i^{(t)}$ is the task history available to $A_i$ under $\mathcal{P}$, $r$ is a relation type, $G_R$ is a relation network, and $\Pi_i(r,G_R)$ is the relational prior inserted into $A_i$'s system prompt. 
Within each matched experimental comparison, we hold the task input $x$, interaction protocol $\mathcal{P}$, history-visibility rule, and decoding settings fixed, varying only $\Pi_i$. The relational prompt can change what agents say to each other, which can then affect their later decisions. 
Relational priors are therefore \emph{protocol-preserving}: they alter the relational context without changing who can speak, which messages are visible, or how final outputs are aggregated.

\subsection{Signed Relation Networks}

A relation network is an undirected graph $G_R=(\mathcal{A},E_R)$ with a sign assignment
\begin{equation}
    s:E_R \rightarrow \{-1,+1\}.
\end{equation}
For an edge $(i,j)\in E_R$, $s_{ij}=+1$ denotes the maximally positive prior for the chosen relation type, while $s_{ij}=-1$ denotes the maximally negative prior. Non-edges correspond to omitted relational priors. The relation network is distinct from the communication graph induced by $\mathcal{P}$: changing $G_R$ changes which pairs receive explicit relational context, not the underlying message-passing protocol.

We study three relation types. \emph{Attitude} specifies how favorably an agent regards another agent \citep{Heider1946AttitudesAC}, \emph{Trust} specifies how much credibility it gives the other agent's statements \citep{Rotter1971GeneralizedEF}, and \emph{Influence} specifies how it responds to the other agent's opinions and proposals \citep{French1956AFT}. The relation type determines what a positive or negative sign means; the topology $E_R$ determines which pairs are explicitly related. When topology is varied, we use representative sparse structures such as chain, star, and tree graphs, illustrated in App.~Fig.~\ref{fig:topology}.

The positivity of a relation network is the fraction of its stated edges assigned a positive sign:
\begin{equation}
    \lambda = \frac{|\{(i,j)\in E_R:s_{ij}=+1\}|}{|E_R|}.
\end{equation}
Thus, $\lambda=0$ is a fully negative explicit relation network, and $\lambda=1$ is a fully positive explicit relation network. For intermediate values, we enumerate all sign assignments consistent with the chosen $\lambda$, covering every placement of positive and negative ties at that positivity.

We also include \neutral{} as a separate control outside the signed $\lambda$-sweep. For this control, every stated edge is assigned $s_{ij}=0$, and $\lambda$ is not defined. In \noprior{}, the relational block is omitted; in \neutral{}, the system prompt explicitly states that every relation is neutral. Both conditions retain the same task role and task instructions. This comparison tests whether explicitly stating neutrality changes behavior relative to omitting relational framing.

The relation network remains fixed throughout each interaction, is symmetric, and is revealed to every agent. The relations are therefore mutual and common knowledge. Every agent sees the entire stated network, including relations between other agents, so all agents have access to the same relational information. In particular, \emph{Influence} specifies mutual responses to each other's opinions and proposals rather than a directed authority relation.

\subsection{Prompt Operationalization}

We implement $\Pi_i(r,G_R)$ by adding a standardized relational block to each agent's system prompt while retaining its task role and task instructions. The added block defines the selected relation type, describes the relation network in natural language, and instructs the agent to consider these relations when making decisions. In the main experiments, the wording template is fixed across conditions except for relation type and edge values. The same signed graph can therefore be instantiated as \emph{Attitude}, \emph{Trust}, or \emph{Influence} while holding the rest of the task prompt constant. App.~Fig.~\ref{box:prompt_template} shows an example GovSim system prompt containing this relational block.
Because the intervention is expressed in natural language, we later test whether the increase in consensus from all-negative to all-positive \emph{Influence} persists under a meaning-preserving paraphrase and a reordering of the relation statements.


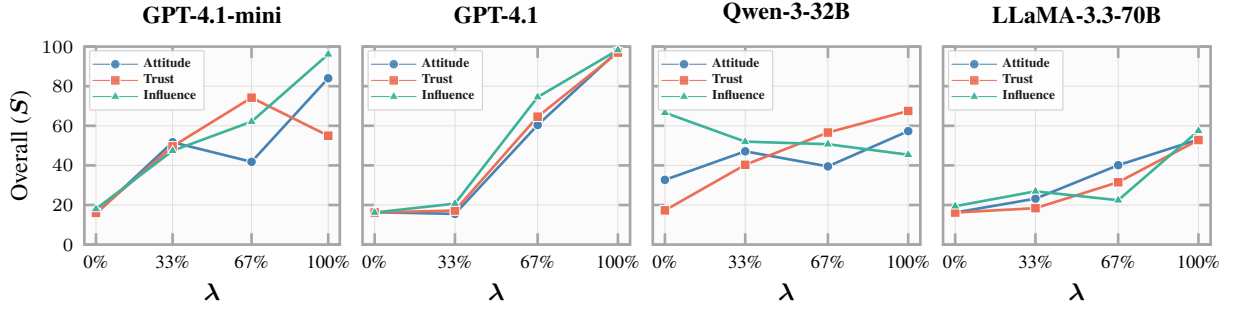
\begin{figure*}[t]
    \centering
    
    \begin{tikzpicture}
        \begin{axis}[
            name=plot1,
            xshift=-2cm,
            width=0.31\textwidth,
            height=4.2cm,
            xlabel={\small $\bm{\lambda}$},
            xlabel style={yshift=2pt},
            ylabel={\small Overall ($\bm{S}$)},
            ylabel style={yshift=-4pt},
            xmin=-5, xmax=105,
            ymin=0, ymax=100,
            xtick={0,33,67,100},
            xticklabels={0\%,33\%,67\%,100\%},
            ytick={0,20,40,60,80,100},
            tick label style={font=\scriptsize},
            label style={font=\scriptsize},
            title={\small\bfseries GPT-4.1-mini},
            title style={yshift=-0.5mm},
            axis background/.style={fill=gray!3},
            axis line style={gray!70},
            grid=major,
            grid style={gray!25, line width=0.3pt},
            line width=1pt,
            every tick/.style={gray!70},
            legend style={
                at={(0.02,0.98)},
                anchor=north west,
                font=\fontsize{5}{4.5}\selectfont\bfseries,
                fill=white,
                fill opacity=0.9,
                draw=gray!40,
                line width=0.2pt,
                inner xsep=1pt,
                inner ysep=0.5pt,
                row sep=-3pt,
                rounded corners=0.5pt,
                /tikz/every even column/.append style={column sep=1pt},
            },
            legend cell align=left,
        ]
        \addplot[color={rgb,255:red,70;green,130;blue,180}, mark=*, mark size=1.8pt, mark options={fill={rgb,255:red,70;green,130;blue,180}, draw=white, line width=0.3pt}] coordinates {
            (0,15.6) (33,51.7) (67,41.8) (100,84.0)
        };
        \addlegendentry{Attitude}
        \addplot[color={rgb,255:red,235;green,110;blue,85}, mark=square*, mark size=1.8pt, mark options={fill={rgb,255:red,235;green,110;blue,85}, draw=white, line width=0.3pt}] coordinates {
            (0,16.0) (33,49.6) (67,74.2) (100,55.0)
        };
        \addlegendentry{Trust}
        \addplot[color={rgb,255:red,60;green,179;blue,153}, mark=triangle*, mark size=2.1pt, mark options={fill={rgb,255:red,60;green,179;blue,153}, draw=white, line width=0.3pt}] coordinates {
            (0,18.0) (33,47.4) (67,62.1) (100,96.1)
        };
        \addlegendentry{Influence}
        \end{axis}
        
        \begin{axis}[
            name=plot2,
            at={($(plot1.east)+(0.3cm,0)$)},
            anchor=west,
            width=0.32\textwidth,
            height=4.2cm,
            xlabel={\small $\bm{\lambda}$},
            xlabel style={yshift=2pt},
            xmin=-5, xmax=105,
            ymin=0, ymax=100,
            xtick={0,33,67,100},
            xticklabels={0\%,33\%,67\%,100\%},
            ytick={0,20,40,60,80,100},
            yticklabels={},
            tick label style={font=\scriptsize},
            label style={font=\scriptsize},
            title={\small\bfseries GPT-4.1},
            title style={yshift=-0.5mm},
            axis background/.style={fill=gray!3},
            axis line style={gray!70},
            grid=major,
            grid style={gray!25, line width=0.3pt},
            line width=1pt,
            every tick/.style={gray!70},
            legend style={
                at={(0.02,0.98)},
                anchor=north west,
                font=\fontsize{5}{4.5}\selectfont\bfseries,
                fill=white,
                fill opacity=0.9,
                draw=gray!40,
                line width=0.2pt,
                inner xsep=1pt,
                inner ysep=0.5pt,
                row sep=-3pt,
                rounded corners=0.5pt,
                /tikz/every even column/.append style={column sep=1pt},
            },
            legend cell align=left,
        ]
        \addplot[color={rgb,255:red,70;green,130;blue,180}, mark=*, mark size=1.8pt, mark options={fill={rgb,255:red,70;green,130;blue,180}, draw=white, line width=0.3pt}] coordinates {
            (0,16.4) (33,15.5) (67,60.4) (100,97.8)
        };
        \addlegendentry{Attitude}
        \addplot[color={rgb,255:red,235;green,110;blue,85}, mark=square*, mark size=1.8pt, mark options={fill={rgb,255:red,235;green,110;blue,85}, draw=white, line width=0.3pt}] coordinates {
            (0,16.2) (33,17.2) (67,64.6) (100,97.0)
        };
        \addlegendentry{Trust}
        \addplot[color={rgb,255:red,60;green,179;blue,153}, mark=triangle*, mark size=2.1pt, mark options={fill={rgb,255:red,60;green,179;blue,153}, draw=white, line width=0.3pt}] coordinates {
            (0,16.2) (33,20.7) (67,74.5) (100,98.3)
        };
        \addlegendentry{Influence}
        \end{axis}
        
        \begin{axis}[
            name=plot3,
            at={($(plot2.east)+(0.3cm,0)$)},
            anchor=west,
            width=0.32\textwidth,
            height=4.2cm,
            xlabel={\small $\bm{\lambda}$},
            xlabel style={yshift=2pt},
            xmin=-5, xmax=105,
            ymin=0, ymax=100,
            xtick={0,33,67,100},
            xticklabels={0\%,33\%,67\%,100\%},
            ytick={0,20,40,60,80,100},
            yticklabels={},
            tick label style={font=\scriptsize},
            label style={font=\scriptsize},
            title={\small\bfseries Qwen-3-32B},
            title style={yshift=-0.5mm},
            axis background/.style={fill=gray!3},
            axis line style={gray!70},
            grid=major,
            grid style={gray!25, line width=0.3pt},
            line width=1pt,
            every tick/.style={gray!70},
            legend style={
                at={(0.02,0.98)},
                anchor=north west,
                font=\fontsize{5}{4.5}\selectfont\bfseries,
                fill=white,
                fill opacity=0.9,
                draw=gray!40,
                line width=0.2pt,
                inner xsep=1pt,
                inner ysep=0.5pt,
                row sep=-3pt,
                rounded corners=0.5pt,
                /tikz/every even column/.append style={column sep=1pt},
            },
            legend cell align=left,
        ]
        \addplot[color={rgb,255:red,70;green,130;blue,180}, mark=*, mark size=1.8pt, mark options={fill={rgb,255:red,70;green,130;blue,180}, draw=white, line width=0.3pt}] coordinates {
            (0,32.7) (33,47.1) (67,39.5) (100,57.3)
        };
        \addlegendentry{Attitude}
        \addplot[color={rgb,255:red,235;green,110;blue,85}, mark=square*, mark size=1.8pt, mark options={fill={rgb,255:red,235;green,110;blue,85}, draw=white, line width=0.3pt}] coordinates {
            (0,17.3) (33,40.3) (67,56.6) (100,67.5)
        };
        \addlegendentry{Trust}
        \addplot[color={rgb,255:red,60;green,179;blue,153}, mark=triangle*, mark size=2.1pt, mark options={fill={rgb,255:red,60;green,179;blue,153}, draw=white, line width=0.3pt}] coordinates {
            (0,66.6) (33,52.0) (67,50.7) (100,45.4)
        };
        \addlegendentry{Influence}
        \end{axis}
        
        \begin{axis}[
            name=plot4,
            at={($(plot3.east)+(0.3cm,0)$)},
            anchor=west,
            width=0.32\textwidth,
            height=4.2cm,
            xlabel={\small $\bm{\lambda}$},
            xlabel style={yshift=2pt},
            xmin=-5, xmax=105,
            ymin=0, ymax=100,
            xtick={0,33,67,100},
            xticklabels={0\%,33\%,67\%,100\%},
            ytick={0,20,40,60,80,100},
            yticklabels={},
            tick label style={font=\scriptsize},
            label style={font=\scriptsize},
            title={\small\bfseries LLaMA-3.3-70B},
            title style={yshift=-0.5mm},
            axis background/.style={fill=gray!3},
            axis line style={gray!70},
            grid=major,
            grid style={gray!25, line width=0.3pt},
            line width=1pt,
            every tick/.style={gray!70},
            legend style={
                at={(0.02,0.98)},
                anchor=north west,
                font=\fontsize{5}{4.5}\selectfont\bfseries,
                fill=white,
                fill opacity=0.9,
                draw=gray!40,
                line width=0.2pt,
                inner xsep=1pt,
                inner ysep=0.5pt,
                row sep=-3pt,
                rounded corners=0.5pt,
                /tikz/every even column/.append style={column sep=1pt},
            },
            legend cell align=left,
        ]
        \addplot[color={rgb,255:red,70;green,130;blue,180}, mark=*, mark size=1.8pt, mark options={fill={rgb,255:red,70;green,130;blue,180}, draw=white, line width=0.3pt}] coordinates {
            (0,16.2) (33,23.2) (67,40.1) (100,53.1)
        };
        \addlegendentry{Attitude}
        \addplot[color={rgb,255:red,235;green,110;blue,85}, mark=square*, mark size=1.8pt, mark options={fill={rgb,255:red,235;green,110;blue,85}, draw=white, line width=0.3pt}] coordinates {
            (0,16.2) (33,18.4) (67,31.5) (100,52.8)
        };
        \addlegendentry{Trust}
        \addplot[color={rgb,255:red,60;green,179;blue,153}, mark=triangle*, mark size=2.1pt, mark options={fill={rgb,255:red,60;green,179;blue,153}, draw=white, line width=0.3pt}] coordinates {
            (0,19.4) (33,26.9) (67,22.4) (100,57.5)
        };
        \addlegendentry{Influence}
        \end{axis}
    \end{tikzpicture}

    \vspace{-0.5em}
    
    \caption{\small Three-agent GovSim sustainability-oriented overall score $S$ (higher is better) across model backbones, relation types, and values of $\lambda$. Scores are generally higher under all-positive than all-negative relations.}
    \label{fig:govsim_3agents_lambas}
\end{figure*}

\section{Experiments}
\label{sec:experiments}

We instantiate relational priors in two task families: GovSim for repeated commons governance and multi-agent debate for subjective and objective questions. The design separates three comparisons. First, the within-prior sweep over $\lambda$ tests how increasing relational positivity changes group dynamics. Second, comparison with \noprior{} tests whether adding an explicit relational block improves the target metric relative to omitting relational framing. Third, comparing \neutral{} with \noprior{} tests whether explicitly stating neutral relations changes group behavior.
Relation positivity is the primary manipulation throughout. We vary relation type and model backbone in both task families. Relation topology varies only in GovSim, where larger teams allow meaningful sparse relation structures; debate uses complete connectivity to study changes in agreement under a fixed interaction structure. For every intermediate $\lambda$, we enumerate all sign assignments consistent with that positivity, covering every placement of positive and negative edges.

\paragraph{Statistical uncertainty.}

We summarize the broader sweeps descriptively and report 95\% bootstrap confidence intervals (CIs) for selected all-positive minus all-negative performance differences. We resample complete simulation runs for GovSim and questions for debate, keeping all agents and months or rounds within each run or question together. For debate, the same sampled questions are used in both conditions. We recalculate the performance difference in each resample and use the resulting distribution to obtain the interval. Apps.~\ref{app:govsim-uncertainty} and~\ref{app:debate-uncertainty} provide the procedures for each task.

\subsection{GovSim}

We use the Governance of the Commons Simulation (GovSim) \citep{piatti2024cooperate} to study repeated cooperation around a shared resource. Agents act as fishermen who repeatedly decide how much to harvest from a renewable common resource, alternating between simultaneous harvest decisions and free-form discussion. Episodes last at most 12 months and terminate early if the resource collapses. Strong performance requires the group to limit its combined harvest so that the resource can replenish, rather than maximize immediate individual extraction. Full environment dynamics are given in App.~\ref{sec:govsim_details_dynamics}.

We evaluate both three-agent and five-agent GovSim settings. Each team uses a single model backbone, holding model capability constant across agents when testing relational priors. This separates the intervention from capability differences that might independently create leadership, deference, or exploitation dynamics. Teams are instantiated separately with GPT-4.1-mini, GPT-4.1, Qwen-3-32B, and LLaMA-3.3-70B-Instruct. The relation network is complete in the three-agent setting; in the five-agent setting, we additionally consider chain, star, and tree topologies. We set temperature to 0.8 and evaluate each fixed model--relation--topology--sign-assignment condition over five complete runs. 


We report the five GovSim metrics from the original benchmark: survival time, survival rate, total gain, inequality, and over-usage. For compact comparison across conditions, we use a normalized weighted aggregate score $S$, with higher values indicating better sustainability-oriented performance. The weights emphasize survival time, survival rate, and reduced over-usage. Exact definitions and weights are provided in App.~\ref{sec:govsim_details_metrics}.

\input{figures/fig03_govsim_delta}

\subsection{Debate}

We next study multi-agent debate, where relational priors can change both whether agents agree and whether their agreement is correct. Each debate contains three answer rounds: agents first answer independently, then complete two revision rounds in which they observe the other agents' responses from the preceding round and may revise their own answers, following the same AutoGen-style debate protocol \citep{wu2024autogen}. This setting lets us separate subjective consensus from objective correctness. Full debate dynamics are described in App.~\ref{sec:debate_formalization}.

Unlike GovSim, debate teams contain different models to introduce differences in initial answers and confidence. Three-agent teams allow both 2--1 majorities and three-way disagreement while keeping the experiments tractable. Each revision round exposes every agent to the other agents' previous-round responses, so the interaction structure is complete. We also use a complete relation network. We evaluate one GPT group (GPT-4.1-nano, GPT-4.1-mini, and GPT-4o-mini) and one open-weight group (LLaMA-3.3-70B-Instruct, Qwen-3-32B, and Mixtral-8x22B-Instruct-v0.1), using greedy decoding throughout.

For subjective debate, we use OpinionQA \citep{pmlr-v202-santurkar23a} and report final-round consensus rate (CR): the fraction of initially disagreeing questions that end in unanimous agreement. For objective debate, we use 1,000 questions from MMLU-Pro \citep{wang2024mmlupro} and all questions from GPQA-Diamond \citep{rein2024gpqa}. We report the same CR, final majority-vote accuracy (ACC) over all evaluated questions, and Consensus Correctness Rate (CCR). CCR is the fraction of consensuses formed from initial disagreement that are correct. Initial disagreement is determined separately for each condition because relational prompts are present from the first answer. ACC therefore evaluates all questions, whereas CR and CCR describe consensus formation among initially disagreeing questions. Formal definitions are given in App.~\ref{sec:debate_evaluation_details}.

\section{Results and Analysis}
\label{sec:results}



\subsection{GovSim: Positivity Supports Sustainable Coordination}


More positive relations generally improve sustainability within the GovSim relational-prior sweep. The aggregate score $S$ is typically higher under all-positive than all-negative relations in both the three-agent setting (Fig.~\ref{fig:govsim_3agents_lambas}) and the five-agent setting (App.~Fig.~\ref{fig:govsim_5agents_lambas}), with reversals at intermediate positivity levels in several sweeps.
For the complete three-agent graph, 10 of the 12 all-positive minus all-negative score differences have 95\% run-level bootstrap CIs entirely above zero. The exceptions are Qwen-3-32B under \emph{Attitude} ($\Delta S=+24.7$, 95\% CI: $[-13.7,+63.4]$) and \emph{Influence} ($\Delta S=-21.1$, 95\% CI: $[-52.3,+15.8]$), where the direction of the difference remains uncertain. Full results are reported in App.~\ref{app:govsim-uncertainty}.

Whether adding fully positive relations improves on \noprior{} is a separate, model-dependent question (Fig.~\ref{fig:govsim_delta_s}). In five-agent GovSim, GPT-4.1 remains close to its no-prior score across relation types and topologies. LLaMA-3.3-70B-Instruct gains substantially in several configurations, while GPT-4.1-mini and Qwen-3-32B show both gains and losses. Thus, higher positivity generally helps relative to negative relations, while gains over omitting relational framing depend on the model and configuration.

\begin{figure}[t]
\centering
\definecolor{attitudecolor}{RGB}{70,130,180}
\definecolor{trustcolor}{RGB}{235,110,85}
\definecolor{influencecolor}{RGB}{60,179,153}
\begin{tikzpicture}
\begin{axis}[
    name=neutralaxis,
    width=\columnwidth,
    height=4.1cm,
    ylabel={\textbf{Overall} ($\bm{S}$)},
    ylabel style={font=\small, yshift=-5pt},
    ymin=0, ymax=110,
    xmin=0.5, xmax=4.5,
    ytick={0,25,50,75,100},
    xtick={1,2,3,4},
    xticklabels={\textbf{GPT-4.1-m}, \textbf{GPT-4.1}, \textbf{Qwen}, \textbf{LLaMA}},
    xticklabel style={font=\scriptsize},
    tick label style={font=\scriptsize},
    axis background/.style={fill=gray!5},
    axis line style={gray!70},
    grid=major,
    grid style={gray!25, line width=0.3pt},
    every tick/.style={gray!70},
    clip=false,
]
\fill[attitudecolor] (axis cs:0.65,0) rectangle (axis cs:0.87,58.9);
\fill[trustcolor] (axis cs:0.89,0) rectangle (axis cs:1.11,55.8);
\fill[influencecolor] (axis cs:1.13,0) rectangle (axis cs:1.35,61.2);
\draw[black, dashed, line width=1pt] (axis cs:0.62,91.0) -- (axis cs:1.38,91.0);
\fill[attitudecolor] (axis cs:1.65,0) rectangle (axis cs:1.87,94.0);
\fill[trustcolor] (axis cs:1.89,0) rectangle (axis cs:2.11,47.5);
\fill[influencecolor] (axis cs:2.13,0) rectangle (axis cs:2.35,65.9);
\draw[black, dashed, line width=1pt] (axis cs:1.62,92.6) -- (axis cs:2.38,92.6);
\fill[attitudecolor] (axis cs:2.65,0) rectangle (axis cs:2.87,56.2);
\fill[trustcolor] (axis cs:2.89,0) rectangle (axis cs:3.11,54.3);
\fill[influencecolor] (axis cs:3.13,0) rectangle (axis cs:3.35,52.2);
\draw[black, dashed, line width=1pt] (axis cs:2.62,66.9) -- (axis cs:3.38,66.9);
\fill[attitudecolor] (axis cs:3.65,0) rectangle (axis cs:3.87,20.2);
\fill[trustcolor] (axis cs:3.89,0) rectangle (axis cs:4.11,17.2);
\fill[influencecolor] (axis cs:4.13,0) rectangle (axis cs:4.35,16.4);
\draw[black, dashed, line width=1pt] (axis cs:3.62,24.0) -- (axis cs:4.38,24.0);
\end{axis}
\node[draw=gray!60, fill=white, rounded corners=2pt, inner xsep=2pt, inner ysep=1pt, line width=0.4pt, anchor=north] at ([yshift=-0.5cm]neutralaxis.south) {\tikz[baseline=-0.5ex]{
    \node[fill=attitudecolor, rounded corners=0.5pt, minimum width=6pt, minimum height=6pt, inner sep=0pt] (c1) {};
    \node[right=1pt of c1, font=\scriptsize] (l1) {Attitude};
    \node[fill=trustcolor, rounded corners=0.5pt, minimum width=6pt, minimum height=6pt, inner sep=0pt, right=3pt of l1] (c2) {};
    \node[right=1pt of c2, font=\scriptsize] (l2) {Trust};
    \node[fill=influencecolor, rounded corners=0.5pt, minimum width=6pt, minimum height=6pt, inner sep=0pt, right=3pt of l2] (c3) {};
    \node[right=1pt of c3, font=\scriptsize] (l3) {Influence};
    \draw[black, dashed, line width=0.8pt] ([xshift=4pt]l3.east) -- ++(0.3cm,0) coordinate (de);
    \node[right=1pt of de, font=\scriptsize] {\noprior{}};
}};
\end{tikzpicture}
\vspace{-0.5em}
\caption{\small Three-agent GovSim overall score $S$ under \neutral{}, by model backbone and relation type. Dashed lines show the corresponding \noprior{} baselines. Model labels shorten GPT-4.1-mini, Qwen-3-32B, and LLaMA-3.3-70B-Instruct to ``GPT-4.1-m'', ``Qwen'', and ``LLaMA'', respectively.}
\label{fig:neutral_baseline_3agents}
\end{figure}
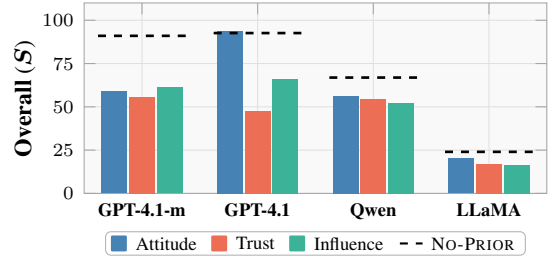

\begin{figure*}[t]
\centering
\definecolor{attitudecolor}{RGB}{70,130,180}
\definecolor{trustcolor}{RGB}{235,110,85}
\definecolor{influencecolor}{RGB}{60,179,153}
\resizebox{\textwidth}{!}{%
\begin{tikzpicture}
\pgfplotsset{debateCR/.style={
    width=3.2cm, height=3.8cm,
    xmin=-5, xmax=105, ymin=-50, ymax=10,
    xtick={0,33,67,100}, xticklabels={0\%,33\%,67\%,100\%},
    ytick={-50,-30,-10,10}, tick label style={font=\scriptsize},
    xticklabel style={font=\tiny}, label style={font=\scriptsize},
    axis background/.style={fill=gray!3}, axis line style={gray!70},
    grid=major, grid style={gray!25, line width=0.3pt},
    line width=1pt, every tick/.style={gray!70}
}}
\begin{axis}[debateCR, name=plot1, ylabel={\scriptsize $\bm{\Delta CR}$}, ylabel style={yshift=-5pt}]
\addplot[gray!50, dashed, line width=0.6pt, forget plot] coordinates {(-5,0) (105,0)}; \node[anchor=south, font=\fontsize{5}{4}\selectfont, text=gray!60, yshift=-1pt] at (axis cs:50,0) {69.5};
\addplot[color=attitudecolor, mark=*, mark size=1.8pt, mark options={fill=attitudecolor, draw=white, line width=0.3pt}] coordinates {(0,-21.8) (33,-20.5) (67,-11.0) (100,-0.4)};
\addplot[color=trustcolor, mark=square*, mark size=1.8pt, mark options={fill=trustcolor, draw=white, line width=0.3pt}] coordinates {(0,-30.5) (33,-24.7) (67,-8.6) (100,-1.2)};
\addplot[color=influencecolor, mark=triangle*, mark size=2.1pt, mark options={fill=influencecolor, draw=white, line width=0.3pt}] coordinates {(0,-44.7) (33,-35.9) (67,-20.6) (100,-14.3)};
\end{axis}
\begin{axis}[debateCR, name=plot2, at={($(plot1.east)+(0.35cm,0)$)}, anchor=west, yticklabels={}]
\addplot[gray!50, dashed, line width=0.6pt, forget plot] coordinates {(-5,0) (105,0)}; \node[anchor=south, font=\fontsize{5}{4}\selectfont, text=gray!60, yshift=-1pt] at (axis cs:50,0) {63.1};
\addplot[color=attitudecolor, mark=*, mark size=1.8pt, mark options={fill=attitudecolor, draw=white, line width=0.3pt}] coordinates {(0,-18.3) (33,-27.4) (67,-11.0) (100,5.8)};
\addplot[color=trustcolor, mark=square*, mark size=1.8pt, mark options={fill=trustcolor, draw=white, line width=0.3pt}] coordinates {(0,-25.1) (33,-25.6) (67,-3.3) (100,5.6)};
\addplot[color=influencecolor, mark=triangle*, mark size=2.1pt, mark options={fill=influencecolor, draw=white, line width=0.3pt}] coordinates {(0,-42.2) (33,-35.0) (67,-8.9) (100,-5.8)};
\end{axis}
\node[anchor=south, font=\scriptsize\bfseries] at ($(plot1.north)!0.5!(plot2.north)+(0,0.45cm)$) {OpinionQA};
\node[anchor=south, font=\scriptsize\bfseries] at ($(plot1.north)+(0,0.1cm)$) {GPT Group};
\node[anchor=south, font=\scriptsize\bfseries] at ($(plot2.north)+(0,0.1cm)$) {Open-Weight Group};

\begin{axis}[debateCR, name=plot3, at={($(plot2.east)+(0.65cm,0)$)}, anchor=west, yticklabels={}]
\addplot[gray!50, dashed, line width=0.6pt, forget plot] coordinates {(-5,0) (105,0)}; \node[anchor=south, font=\fontsize{5}{4}\selectfont, text=gray!60, yshift=-1pt] at (axis cs:50,0) {79.1};
\addplot[color=attitudecolor, mark=*, mark size=1.8pt, mark options={fill=attitudecolor, draw=white, line width=0.3pt}] coordinates {(0,-3.1) (33,-1.5) (67,-2.4) (100,-3.2)};
\addplot[color=trustcolor, mark=square*, mark size=1.8pt, mark options={fill=trustcolor, draw=white, line width=0.3pt}] coordinates {(0,-2.7) (33,-8.6) (67,-6.6) (100,-7.6)};
\addplot[color=influencecolor, mark=triangle*, mark size=2.1pt, mark options={fill=influencecolor, draw=white, line width=0.3pt}] coordinates {(0,-8.9) (33,-12.6) (67,-11.1) (100,-6.5)};
\end{axis}
\begin{axis}[debateCR, name=plot4, at={($(plot3.east)+(0.35cm,0)$)}, anchor=west, yticklabels={}]
\addplot[gray!50, dashed, line width=0.6pt, forget plot] coordinates {(-5,0) (105,0)}; \node[anchor=south, font=\fontsize{5}{4}\selectfont, text=gray!60, yshift=-1pt] at (axis cs:50,0) {74.1};
\addplot[color=attitudecolor, mark=*, mark size=1.8pt, mark options={fill=attitudecolor, draw=white, line width=0.3pt}] coordinates {(0,-30.2) (33,-19.8) (67,-11.5) (100,-2.8)};
\addplot[color=trustcolor, mark=square*, mark size=1.8pt, mark options={fill=trustcolor, draw=white, line width=0.3pt}] coordinates {(0,-35.7) (33,-22.2) (67,-8.0) (100,4.8)};
\addplot[color=influencecolor, mark=triangle*, mark size=2.1pt, mark options={fill=influencecolor, draw=white, line width=0.3pt}] coordinates {(0,-40.3) (33,-27.0) (67,-13.9) (100,2.8)};
\end{axis}
\node[anchor=south, font=\scriptsize\bfseries] at ($(plot3.north)!0.5!(plot4.north)+(0,0.45cm)$) {MMLU-Pro};
\node[anchor=south, font=\scriptsize\bfseries] at ($(plot3.north)+(0,0.1cm)$) {GPT Group};
\node[anchor=south, font=\scriptsize\bfseries] at ($(plot4.north)+(0,0.1cm)$) {Open-Weight Group};

\begin{axis}[debateCR, name=plot5, at={($(plot4.east)+(0.65cm,0)$)}, anchor=west, yticklabels={}]
\addplot[gray!50, dashed, line width=0.6pt, forget plot] coordinates {(-5,0) (105,0)}; \node[anchor=south, font=\fontsize{5}{4}\selectfont, text=gray!60, yshift=-1pt] at (axis cs:50,0) {80.6};
\addplot[color=attitudecolor, mark=*, mark size=1.8pt, mark options={fill=attitudecolor, draw=white, line width=0.3pt}] coordinates {(0,0.5) (33,-5.3) (67,-3.9) (100,-6.4)};
\addplot[color=trustcolor, mark=square*, mark size=1.8pt, mark options={fill=trustcolor, draw=white, line width=0.3pt}] coordinates {(0,3.1) (33,-1.2) (67,-9.2) (100,0.2)};
\addplot[color=influencecolor, mark=triangle*, mark size=2.1pt, mark options={fill=influencecolor, draw=white, line width=0.3pt}] coordinates {(0,-8.3) (33,-10.8) (67,-13.0) (100,-10.6)};
\end{axis}
\begin{axis}[debateCR, name=plot6, at={($(plot5.east)+(0.35cm,0)$)}, anchor=west, yticklabels={}, legend to name=sharedlegend, legend style={font=\fontsize{6}{5}\selectfont\bfseries, fill=white, draw=gray!40, line width=0.3pt, inner xsep=2pt, inner ysep=1pt, rounded corners=1pt, legend columns=3, column sep=3pt}, legend cell align=left]
\addplot[gray!50, dashed, line width=0.6pt, forget plot] coordinates {(-5,0) (105,0)}; \node[anchor=south, font=\fontsize{5}{4}\selectfont, text=gray!60, yshift=-1pt] at (axis cs:50,0) {81.2};
\addplot[color=attitudecolor, mark=*, mark size=1.5pt, mark options={fill=attitudecolor, draw=white, line width=0.3pt}] coordinates {(0,-22.8) (33,-13.4) (67,-13.6) (100,-7.1)}; \addlegendentry{\textcolor{attitudecolor}{Attitude}}
\addplot[color=trustcolor, mark=square*, mark size=1.5pt, mark options={fill=trustcolor, draw=white, line width=0.3pt}] coordinates {(0,-20.8) (33,-17.2) (67,-11.8) (100,-8.7)}; \addlegendentry{\textcolor{trustcolor}{Trust}}
\addplot[color=influencecolor, mark=triangle*, mark size=1.8pt, mark options={fill=influencecolor, draw=white, line width=0.3pt}] coordinates {(0,-32.3) (33,-25.5) (67,-13.5) (100,-8.5)}; \addlegendentry{\textcolor{influencecolor}{Influence}}
\end{axis}
\node[anchor=south, font=\scriptsize\bfseries] at ($(plot5.north)!0.5!(plot6.north)+(0,0.45cm)$) {GPQA-Diamond};
\node[anchor=south, font=\scriptsize\bfseries] at ($(plot5.north)+(0,0.1cm)$) {GPT Group};
\node[anchor=south, font=\scriptsize\bfseries] at ($(plot6.north)+(0,0.1cm)$) {Open-Weight Group};
\node[anchor=north, font=\scriptsize] at ($(plot1.south)!0.5!(plot6.south)-(0,0.3cm)$) {$\bm{\lambda}$};
\node[anchor=north] at ($(plot1.south)!0.5!(plot6.south)-(0,0.6cm)$) {\pgfplotslegendfromname{sharedlegend}};
\end{tikzpicture}%
}
\vspace{-2em}
\caption{\small Debate consensus rate (CR) differences in percentage points relative to \noprior{}. On OpinionQA, fully positive priors produce more consensus than fully negative priors in all six comparisons. Dashed lines mark zero difference; annotations show \noprior{} CR.}
\label{fig:cr_comparison}
\end{figure*}
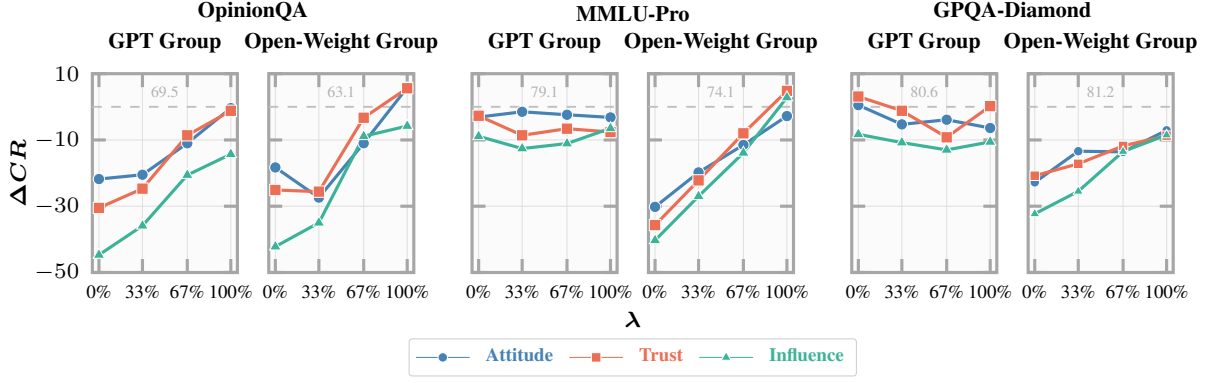

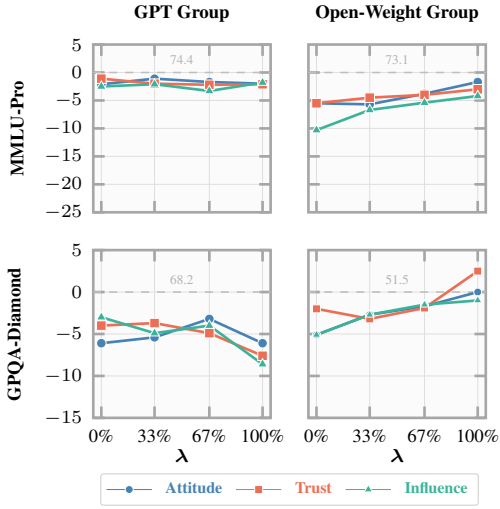
\begin{figure}[t]
\centering
\definecolor{attitudecolor}{RGB}{70,130,180}
\definecolor{trustcolor}{RGB}{235,110,85}
\definecolor{influencecolor}{RGB}{60,179,153}
\begin{tikzpicture}
\pgfplotsset{accaxis/.style={
    width=0.51\columnwidth, height=3.8cm,
    xmin=-5, xmax=105, tick label style={font=\scriptsize}, label style={font=\scriptsize},
    axis background/.style={fill=gray!3}, axis line style={gray!70}, grid=major,
    grid style={gray!25, line width=0.3pt}, line width=1pt, every tick/.style={gray!70}
}}
\begin{axis}[accaxis, name=plot1, ylabel={\scriptsize\bfseries MMLU-Pro}, ymin=-25, ymax=5, xtick={0,33,67,100}, xticklabels={}, ytick={-25,-20,-15,-10,-5,0,5}, title={\scriptsize\bfseries GPT Group}, title style={yshift=-0.5mm}]
\addplot[gray!50, dashed, line width=0.6pt, forget plot] coordinates {(-5,0) (105,0)}; \node[anchor=south, font=\fontsize{5}{4}\selectfont, text=gray!60, yshift=-1pt] at (axis cs:50,0) {74.4};
\addplot[color=attitudecolor, mark=*, mark size=1.8pt, mark options={fill=attitudecolor, draw=white, line width=0.3pt}] coordinates {(0,-2.1) (33,-1.1) (67,-1.7) (100,-2.0)};
\addplot[color=trustcolor, mark=square*, mark size=1.8pt, mark options={fill=trustcolor, draw=white, line width=0.3pt}] coordinates {(0,-1.1) (33,-2.0) (67,-2.2) (100,-2.1)};
\addplot[color=influencecolor, mark=triangle*, mark size=2.1pt, mark options={fill=influencecolor, draw=white, line width=0.3pt}] coordinates {(0,-2.5) (33,-2.1) (67,-3.3) (100,-1.8)};
\end{axis}
\begin{axis}[accaxis, name=plot2, at={($(plot1.east)+(0.5cm,0)$)}, anchor=west, ymin=-25, ymax=5, xtick={0,33,67,100}, xticklabels={}, ytick={-20,-15,-10,-5,0}, yticklabels={}, title={\scriptsize\bfseries Open-Weight Group}, title style={yshift=-0.5mm}]
\addplot[gray!50, dashed, line width=0.6pt, forget plot] coordinates {(-5,0) (105,0)}; \node[anchor=south, font=\fontsize{5}{4}\selectfont, text=gray!60, yshift=-1pt] at (axis cs:50,0) {73.1};
\addplot[color=attitudecolor, mark=*, mark size=1.8pt, mark options={fill=attitudecolor, draw=white, line width=0.3pt}] coordinates {(0,-5.5) (33,-5.7) (67,-3.8) (100,-1.7)};
\addplot[color=trustcolor, mark=square*, mark size=1.8pt, mark options={fill=trustcolor, draw=white, line width=0.3pt}] coordinates {(0,-5.5) (33,-4.5) (67,-4.0) (100,-3.0)};
\addplot[color=influencecolor, mark=triangle*, mark size=2.1pt, mark options={fill=influencecolor, draw=white, line width=0.3pt}] coordinates {(0,-10.3) (33,-6.7) (67,-5.4) (100,-4.2)};
\end{axis}
\begin{axis}[accaxis, name=plot3, at={($(plot1.south)-(0,0.5cm)$)}, anchor=north, xlabel={\scriptsize $\bm{\lambda}$}, xlabel style={yshift=5pt}, ylabel={\scriptsize\bfseries GPQA-Diamond}, ymin=-15, ymax=5, xtick={0,33,67,100}, xticklabels={0\%,33\%,67\%,100\%}, ytick={-15,-10,-5,0,5}]
\addplot[gray!50, dashed, line width=0.6pt, forget plot] coordinates {(-5,0) (105,0)}; \node[anchor=south, font=\fontsize{5}{4}\selectfont, text=gray!60, yshift=-1pt] at (axis cs:50,0) {68.2};
\addplot[color=attitudecolor, mark=*, mark size=1.8pt, mark options={fill=attitudecolor, draw=white, line width=0.3pt}] coordinates {(0,-6.1) (33,-5.4) (67,-3.2) (100,-6.1)};
\addplot[color=trustcolor, mark=square*, mark size=1.8pt, mark options={fill=trustcolor, draw=white, line width=0.3pt}] coordinates {(0,-4.0) (33,-3.7) (67,-4.9) (100,-7.6)};
\addplot[color=influencecolor, mark=triangle*, mark size=2.1pt, mark options={fill=influencecolor, draw=white, line width=0.3pt}] coordinates {(0,-3.0) (33,-4.9) (67,-4.0) (100,-8.6)};
\end{axis}
\begin{axis}[accaxis, name=plot4, at={($(plot3.east)+(0.5cm,0)$)}, anchor=west, xlabel={\scriptsize $\bm{\lambda}$}, xlabel style={yshift=5pt}, ymin=-15, ymax=5, xtick={0,33,67,100}, xticklabels={0\%,33\%,67\%,100\%}, ytick={-15,-10,-5,0}, yticklabels={}, legend to name=sharedlegendA, legend style={font=\fontsize{6}{5}\selectfont\bfseries, fill=white, draw=gray!40, line width=0.3pt, inner xsep=2pt, inner ysep=1pt, rounded corners=1pt, legend columns=3, column sep=3pt}, legend cell align=left]
\addplot[gray!50, dashed, line width=0.6pt, forget plot] coordinates {(-5,0) (105,0)}; \node[anchor=south, font=\fontsize{5}{4}\selectfont, text=gray!60, yshift=-1pt] at (axis cs:50,0) {51.5};
\addplot[color=attitudecolor, mark=*, mark size=1.5pt, mark options={fill=attitudecolor, draw=white, line width=0.3pt}] coordinates {(0,-5.1) (33,-2.7) (67,-1.7) (100,0.0)}; \addlegendentry{\textcolor{attitudecolor}{Attitude}}
\addplot[color=trustcolor, mark=square*, mark size=1.5pt, mark options={fill=trustcolor, draw=white, line width=0.3pt}] coordinates {(0,-2.0) (33,-3.2) (67,-1.9) (100,2.5)}; \addlegendentry{\textcolor{trustcolor}{Trust}}
\addplot[color=influencecolor, mark=triangle*, mark size=1.8pt, mark options={fill=influencecolor, draw=white, line width=0.3pt}] coordinates {(0,-5.1) (33,-2.7) (67,-1.5) (100,-1.0)}; \addlegendentry{\textcolor{influencecolor}{Influence}}
\end{axis}
\node[anchor=north] at ($(plot3.south)!0.5!(plot4.south)-(0,0.6cm)$) {\pgfplotslegendfromname{sharedlegendA}};
\end{tikzpicture}
\vspace{-0.5em}
\caption{\small Debate accuracy (ACC) differences in percentage points relative to \noprior{}. Fully positive priors remain below \noprior{} accuracy in 10 of 12 comparisons. Dashed lines mark zero difference; annotations show \noprior{} ACC.}
\label{fig:acc_comparison}
\end{figure}
\begin{figure}[t]
\centering
\definecolor{attitudecolor}{RGB}{70,130,180}
\definecolor{trustcolor}{RGB}{235,110,85}
\definecolor{influencecolor}{RGB}{60,179,153}
\begin{tikzpicture}
\pgfplotsset{ccraxis/.style={
    width=0.51\columnwidth, height=3.8cm,
    xmin=-5, xmax=105, tick label style={font=\scriptsize}, label style={font=\scriptsize},
    axis background/.style={fill=gray!3}, axis line style={gray!70}, grid=major,
    grid style={gray!25, line width=0.3pt}, line width=1pt, every tick/.style={gray!70}
}}
\begin{axis}[ccraxis, name=plot1, ylabel={\scriptsize\bfseries MMLU-Pro}, ymin=-10, ymax=10, xtick={0,33,67,100}, xticklabels={}, ytick={-10,-5,0,5,10}, title={\scriptsize\bfseries GPT Group}, title style={yshift=-0.5mm}]
\addplot[gray!50, dashed, line width=0.6pt, forget plot] coordinates {(-5,0) (105,0)}; \node[anchor=south, font=\fontsize{5}{4}\selectfont, text=gray!60, yshift=-1pt] at (axis cs:50,0) {63.5};
\addplot[color=attitudecolor, mark=*, mark size=1.8pt, mark options={fill=attitudecolor, draw=white, line width=0.3pt}] coordinates {(0,0.2) (33,-1.1) (67,-1.1) (100,-0.5)};
\addplot[color=trustcolor, mark=square*, mark size=1.8pt, mark options={fill=trustcolor, draw=white, line width=0.3pt}] coordinates {(0,1.1) (33,0.5) (67,-0.9) (100,0.3)};
\addplot[color=influencecolor, mark=triangle*, mark size=2.1pt, mark options={fill=influencecolor, draw=white, line width=0.3pt}] coordinates {(0,2.1) (33,0.4) (67,-2.3) (100,-1.2)};
\end{axis}
\begin{axis}[ccraxis, name=plot2, at={($(plot1.east)+(0.5cm,0)$)}, anchor=west, ymin=-10, ymax=10, xtick={0,33,67,100}, xticklabels={}, ytick={-10,-5,0,5,10}, yticklabels={}, title={\scriptsize\bfseries Open-Weight Group}, title style={yshift=-0.5mm}]
\addplot[gray!50, dashed, line width=0.6pt, forget plot] coordinates {(-5,0) (105,0)}; \node[anchor=south, font=\fontsize{5}{4}\selectfont, text=gray!60, yshift=-1pt] at (axis cs:50,0) {65.4};
\addplot[color=attitudecolor, mark=*, mark size=1.8pt, mark options={fill=attitudecolor, draw=white, line width=0.3pt}] coordinates {(0,1.3) (33,0.1) (67,-4.1) (100,-1.2)};
\addplot[color=trustcolor, mark=square*, mark size=1.8pt, mark options={fill=trustcolor, draw=white, line width=0.3pt}] coordinates {(0,7.4) (33,-0.4) (67,-3.6) (100,-2.5)};
\addplot[color=influencecolor, mark=triangle*, mark size=2.1pt, mark options={fill=influencecolor, draw=white, line width=0.3pt}] coordinates {(0,1.5) (33,-3.7) (67,-6.0) (100,-7.0)};
\end{axis}
\begin{axis}[ccraxis, name=plot3, at={($(plot1.south)-(0,0.5cm)$)}, anchor=north, xlabel={\scriptsize $\bm{\lambda}$}, xlabel style={yshift=5pt}, ylabel={\scriptsize\bfseries GPQA-Diamond}, ymin=-12, ymax=8, xtick={0,33,67,100}, xticklabels={0\%,33\%,67\%,100\%}, ytick={-12,-8,-4,0,4,8}]
\addplot[gray!50, dashed, line width=0.6pt, forget plot] coordinates {(-5,0) (105,0)}; \node[anchor=south, font=\fontsize{5}{4}\selectfont, text=gray!60, yshift=-1pt] at (axis cs:50,0) {67.6};
\addplot[color=attitudecolor, mark=*, mark size=1.8pt, mark options={fill=attitudecolor, draw=white, line width=0.3pt}] coordinates {(0,-6.9) (33,-5.0) (67,-2.4) (100,-6.4)};
\addplot[color=trustcolor, mark=square*, mark size=1.8pt, mark options={fill=trustcolor, draw=white, line width=0.3pt}] coordinates {(0,-9.3) (33,-1.0) (67,-4.8) (100,-5.7)};
\addplot[color=influencecolor, mark=triangle*, mark size=2.1pt, mark options={fill=influencecolor, draw=white, line width=0.3pt}] coordinates {(0,3.3) (33,-2.5) (67,-1.1) (100,-9.4)};
\end{axis}
\begin{axis}[ccraxis, name=plot4, at={($(plot3.east)+(0.5cm,0)$)}, anchor=west, xlabel={\scriptsize $\bm{\lambda}$}, xlabel style={yshift=5pt}, ymin=-12, ymax=8, xtick={0,33,67,100}, xticklabels={0\%,33\%,67\%,100\%}, ytick={-12,-8,-4,0,4,8}, yticklabels={}, legend to name=sharedlegendB, legend style={font=\fontsize{6}{5}\selectfont\bfseries, fill=white, draw=gray!40, line width=0.3pt, inner xsep=2pt, inner ysep=1pt, rounded corners=1pt, legend columns=3, column sep=3pt}, legend cell align=left]
\addplot[gray!50, dashed, line width=0.6pt, forget plot] coordinates {(-5,0) (105,0)}; \node[anchor=south, font=\fontsize{5}{4}\selectfont, text=gray!60, yshift=-1pt] at (axis cs:50,0) {45.5};
\addplot[color=attitudecolor, mark=*, mark size=1.5pt, mark options={fill=attitudecolor, draw=white, line width=0.3pt}] coordinates {(0,-4.3) (33,-2.5) (67,0.1) (100,2.1)}; \addlegendentry{\textcolor{attitudecolor}{Attitude}}
\addplot[color=trustcolor, mark=square*, mark size=1.5pt, mark options={fill=trustcolor, draw=white, line width=0.3pt}] coordinates {(0,0.9) (33,2.9) (67,1.8) (100,2.5)}; \addlegendentry{\textcolor{trustcolor}{Trust}}
\addplot[color=influencecolor, mark=triangle*, mark size=1.8pt, mark options={fill=influencecolor, draw=white, line width=0.3pt}] coordinates {(0,4.5) (33,-1.6) (67,-1.2) (100,-1.7)}; \addlegendentry{\textcolor{influencecolor}{Influence}}
\end{axis}
\node[anchor=north] at ($(plot3.south)!0.5!(plot4.south)-(0,0.6cm)$) {\pgfplotslegendfromname{sharedlegendB}};
\end{tikzpicture}
\vspace{-0.5em}
\caption{\small Debate Consensus Correctness Rate (CCR) differences in percentage points relative to \noprior{}. On MMLU-Pro, CCR is lower under all-positive than all-negative relations in all six comparisons. Dashed lines mark zero difference; annotations show \noprior{} CCR.}
\label{fig:consensus_with_correction}
\end{figure}
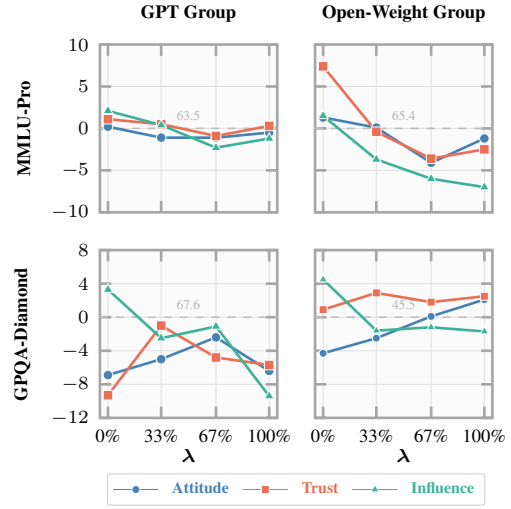

The gains and losses also depend on relation type and topology. Across the fully positive configurations in Fig.~\ref{fig:govsim_delta_s}, \emph{Attitude} has a smaller worst-case loss than \emph{Trust} or \emph{Influence}, while \emph{Influence} spans the widest range of gains and losses. Among the five-agent topologies, chain spans the largest range across models and relation types, with narrower ranges for star and tree. For example, with LLaMA-3.3-70B-Instruct and fully positive \emph{Influence}, the gain over \noprior{} is 41.0 score points for chain, 27.2 for star, and 0.3 for tree. The chain configuration gives the largest observed gain in the figure. These differences show that changing which pairs receive relational cues can substantially change outcomes even when message routing is fixed.

Explicit neutrality also changes outcomes. In three-agent GovSim, \neutral{} scores are below \noprior{} in 11 of the 12 model--relation comparisons; GPT-4.1 under \emph{Attitude} is the exception (Fig.~\ref{fig:neutral_baseline_3agents}). Five-agent neutral controls are also mostly below \noprior{} (App.~Fig.~\ref{fig:neutral_baseline_5agents}). Stating neutral relations is therefore a distinct intervention rather than an interchangeable no-prior control.

\subsection{Debate: Agreement Can Diverge from Correctness}
\label{sec:debate_results}

On OpinionQA, fully positive relations produce more consensus than fully negative relations for both model groups and all three relation types among questions with initial disagreement (left panels of Fig.~\ref{fig:cr_comparison}). Relative to \noprior{}, fully positive \emph{Attitude} and \emph{Trust} improve CR for the open-weight team, while its \emph{Influence} condition and all three GPT conditions remain below baseline. OpinionQA has no universally correct answers, so these results measure agreement; higher CR is useful when consensus itself is the task objective.

\begin{table*}[t]
\centering
\small
\renewcommand{\arraystretch}{1.1}
\setlength{\tabcolsep}{5pt}
\begin{tabular}{@{}lrrr@{}}
\toprule
Initial answers $\rightarrow$ final outcome
& $\lambda=0$
& $\lambda=1$
& $\Delta$ [95\% CI] \\
\midrule
Wrong 2--1 majority $\rightarrow$ correct consensus
& 4.4\% & 4.8\% & $+0.4\;[-4.5,+5.1]$ \\
\addlinespace[3pt]
Correct 2--1 majority $\rightarrow$ wrong consensus
& 2.8\% & 1.9\% & $-0.9\;[-3.5,+1.6]$ \\
\addlinespace[3pt]
Any initial disagreement $\rightarrow$ wrong consensus
& 11.2\% & 32.0\% & $+20.8\;[+16.3,+25.3]$ \\
\bottomrule
\end{tabular}
\caption{Initial-to-final answer transitions for the open-weight MMLU-Pro team under \emph{Influence}. Each percentage is calculated among questions with the initial answer pattern specified in that row, separately for each condition. A 2--1 split means two agents choose the same option and the third chooses a different option; consensus requires all three to agree. The final row measures Wrong Consensus Formation (WCF) and includes both 2--1 splits and cases where all three initial answers differ. Changes are all-positive minus all-negative, in percentage points, with paired 95\% bootstrap confidence intervals.}
\label{tab:answer-transitions}
\end{table*}

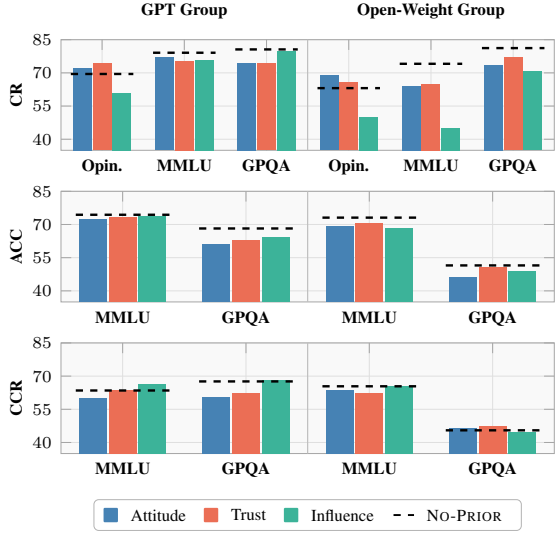
\begin{figure}[t]
\centering
\resizebox{0.95\columnwidth}{!}{%
\begin{tikzpicture}
\definecolor{attitudecolor}{RGB}{70,130,180}
\definecolor{trustcolor}{RGB}{235,110,85}
\definecolor{influencecolor}{RGB}{60,179,153}

\begin{axis}[
    name=crplot,
    width=0.55\textwidth,
    height=3.2cm,
    ylabel={\textbf{CR}},
    ylabel style={font=\scriptsize, at={(axis description cs:-0.06,0.5)}},
    ymin=35, ymax=85,
    xmin=0.5, xmax=6.5,
    ytick={40,55,70,85},
    xtick={1,2,3,4,5,6},
    xticklabels={\textbf{Opin.}, \textbf{MMLU}, \textbf{GPQA}, \textbf{Opin.}, \textbf{MMLU}, \textbf{GPQA}},
    xticklabel style={font=\scriptsize},
    tick label style={font=\scriptsize},
    axis background/.style={fill=gray!5},
    axis line style={gray!70},
    grid=major,
    grid style={gray!25, line width=0.3pt},
    every tick/.style={gray!70},
    clip=false,
]
\fill[attitudecolor] (axis cs:0.65,35) rectangle (axis cs:0.87,72.2);
\fill[trustcolor] (axis cs:0.89,35) rectangle (axis cs:1.11,74.2);
\fill[influencecolor] (axis cs:1.13,35) rectangle (axis cs:1.35,60.9);
\draw[black, dashed, line width=1pt] (axis cs:0.62,69.5) -- (axis cs:1.38,69.5);
\fill[attitudecolor] (axis cs:1.65,35) rectangle (axis cs:1.87,77.0);
\fill[trustcolor] (axis cs:1.89,35) rectangle (axis cs:2.11,75.5);
\fill[influencecolor] (axis cs:2.13,35) rectangle (axis cs:2.35,75.9);
\draw[black, dashed, line width=1pt] (axis cs:1.62,79.1) -- (axis cs:2.38,79.1);
\fill[attitudecolor] (axis cs:2.65,35) rectangle (axis cs:2.87,74.4);
\fill[trustcolor] (axis cs:2.89,35) rectangle (axis cs:3.11,74.2);
\fill[influencecolor] (axis cs:3.13,35) rectangle (axis cs:3.35,80.0);
\draw[black, dashed, line width=1pt] (axis cs:2.62,80.6) -- (axis cs:3.38,80.6);
\fill[attitudecolor] (axis cs:3.65,35) rectangle (axis cs:3.87,68.8);
\fill[trustcolor] (axis cs:3.89,35) rectangle (axis cs:4.11,65.6);
\fill[influencecolor] (axis cs:4.13,35) rectangle (axis cs:4.35,50.0);
\draw[black, dashed, line width=1pt] (axis cs:3.62,63.1) -- (axis cs:4.38,63.1);
\fill[attitudecolor] (axis cs:4.65,35) rectangle (axis cs:4.87,64.1);
\fill[trustcolor] (axis cs:4.89,35) rectangle (axis cs:5.11,64.9);
\fill[influencecolor] (axis cs:5.13,35) rectangle (axis cs:5.35,45.0);
\draw[black, dashed, line width=1pt] (axis cs:4.62,74.1) -- (axis cs:5.38,74.1);
\fill[attitudecolor] (axis cs:5.65,35) rectangle (axis cs:5.87,73.6);
\fill[trustcolor] (axis cs:5.89,35) rectangle (axis cs:6.11,76.9);
\fill[influencecolor] (axis cs:6.13,35) rectangle (axis cs:6.35,70.6);
\draw[black, dashed, line width=1pt] (axis cs:5.62,81.2) -- (axis cs:6.38,81.2);
\node[font=\scriptsize\bfseries] at (axis cs:2,98) {GPT Group};
\node[font=\scriptsize\bfseries] at (axis cs:5,98) {Open-Weight Group};
\draw[gray!50, line width=0.5pt] (axis cs:3.5,35) -- (axis cs:3.5,85);
\end{axis}

\begin{axis}[
    name=accplot,
    at={(crplot.south west)},
    anchor=north west,
    yshift=-0.6cm,
    width=0.55\textwidth,
    height=3.2cm,
    ylabel={\textbf{ACC}},
    ylabel style={font=\scriptsize, at={(axis description cs:-0.06,0.5)}},
    ymin=35, ymax=85,
    xmin=0.5, xmax=4.5,
    ytick={40,55,70,85},
    xtick={1,2,3,4},
    xticklabels={\textbf{MMLU}, \textbf{GPQA}, \textbf{MMLU}, \textbf{GPQA}},
    xticklabel style={font=\scriptsize},
    tick label style={font=\scriptsize},
    axis background/.style={fill=gray!5},
    axis line style={gray!70},
    grid=major,
    grid style={gray!25, line width=0.3pt},
    every tick/.style={gray!70},
    clip=false,
]
\fill[attitudecolor] (axis cs:0.65,35) rectangle (axis cs:0.87,72.4);
\fill[trustcolor] (axis cs:0.89,35) rectangle (axis cs:1.11,73.2);
\fill[influencecolor] (axis cs:1.13,35) rectangle (axis cs:1.35,73.6);
\draw[black, dashed, line width=1pt] (axis cs:0.62,74.4) -- (axis cs:1.38,74.4);
\fill[attitudecolor] (axis cs:1.65,35) rectangle (axis cs:1.87,61.1);
\fill[trustcolor] (axis cs:1.89,35) rectangle (axis cs:2.11,62.6);
\fill[influencecolor] (axis cs:2.13,35) rectangle (axis cs:2.35,64.1);
\draw[black, dashed, line width=1pt] (axis cs:1.62,68.2) -- (axis cs:2.38,68.2);
\fill[attitudecolor] (axis cs:2.65,35) rectangle (axis cs:2.87,69.3);
\fill[trustcolor] (axis cs:2.89,35) rectangle (axis cs:3.11,70.3);
\fill[influencecolor] (axis cs:3.13,35) rectangle (axis cs:3.35,68.1);
\draw[black, dashed, line width=1pt] (axis cs:2.62,73.1) -- (axis cs:3.38,73.1);
\fill[attitudecolor] (axis cs:3.65,35) rectangle (axis cs:3.87,46.0);
\fill[trustcolor] (axis cs:3.89,35) rectangle (axis cs:4.11,50.5);
\fill[influencecolor] (axis cs:4.13,35) rectangle (axis cs:4.35,49.0);
\draw[black, dashed, line width=1pt] (axis cs:3.62,51.5) -- (axis cs:4.38,51.5);
\draw[gray!50, line width=0.5pt] (axis cs:2.5,35) -- (axis cs:2.5,85);
\end{axis}

\begin{axis}[
    name=ccrplot,
    at={(accplot.south west)},
    anchor=north west,
    yshift=-0.6cm,
    width=0.55\textwidth,
    height=3.2cm,
    ylabel={\textbf{CCR}},
    ylabel style={font=\scriptsize, at={(axis description cs:-0.06,0.5)}},
    ymin=35, ymax=85,
    xmin=0.5, xmax=4.5,
    ytick={40,55,70,85},
    xtick={1,2,3,4},
    xticklabels={\textbf{MMLU}, \textbf{GPQA}, \textbf{MMLU}, \textbf{GPQA}},
    xticklabel style={font=\scriptsize},
    tick label style={font=\scriptsize},
    axis background/.style={fill=gray!5},
    axis line style={gray!70},
    grid=major,
    grid style={gray!25, line width=0.3pt},
    every tick/.style={gray!70},
    clip=false,
]
\fill[attitudecolor] (axis cs:0.65,35) rectangle (axis cs:0.87,60.3);
\fill[trustcolor] (axis cs:0.89,35) rectangle (axis cs:1.11,63.9);
\fill[influencecolor] (axis cs:1.13,35) rectangle (axis cs:1.35,66.6);
\draw[black, dashed, line width=1pt] (axis cs:0.62,63.5) -- (axis cs:1.38,63.5);
\fill[attitudecolor] (axis cs:1.65,35) rectangle (axis cs:1.87,60.4);
\fill[trustcolor] (axis cs:1.89,35) rectangle (axis cs:2.11,62.2);
\fill[influencecolor] (axis cs:2.13,35) rectangle (axis cs:2.35,68.0);
\draw[black, dashed, line width=1pt] (axis cs:1.62,67.6) -- (axis cs:2.38,67.6);
\fill[attitudecolor] (axis cs:2.65,35) rectangle (axis cs:2.87,63.6);
\fill[trustcolor] (axis cs:2.89,35) rectangle (axis cs:3.11,62.3);
\fill[influencecolor] (axis cs:3.13,35) rectangle (axis cs:3.35,65.6);
\draw[black, dashed, line width=1pt] (axis cs:2.62,65.4) -- (axis cs:3.38,65.4);
\fill[attitudecolor] (axis cs:3.65,35) rectangle (axis cs:3.87,46.6);
\fill[trustcolor] (axis cs:3.89,35) rectangle (axis cs:4.11,47.6);
\fill[influencecolor] (axis cs:4.13,35) rectangle (axis cs:4.35,44.6);
\draw[black, dashed, line width=1pt] (axis cs:3.62,45.5) -- (axis cs:4.38,45.5);
\draw[gray!50, line width=0.5pt] (axis cs:2.5,35) -- (axis cs:2.5,85);
\end{axis}

\node[draw=gray!60, fill=white, rounded corners=2pt, inner xsep=4pt, inner ysep=2pt, line width=0.4pt] at ([yshift=-0.9cm]ccrplot.south) {\tikz[baseline=-0.5ex]{
    \node[fill=attitudecolor, rounded corners=0.5pt, minimum width=6pt, minimum height=6pt, inner sep=0pt] (c1) {};
    \node[right=1pt of c1, font=\scriptsize] (l1) {Attitude};
    \node[fill=trustcolor, rounded corners=0.5pt, minimum width=6pt, minimum height=6pt, inner sep=0pt, right=3pt of l1] (c2) {};
    \node[right=1pt of c2, font=\scriptsize] (l2) {Trust};
    \node[fill=influencecolor, rounded corners=0.5pt, minimum width=6pt, minimum height=6pt, inner sep=0pt, right=3pt of l2] (c3) {};
    \node[right=1pt of c3, font=\scriptsize] (l3) {Influence};
    \draw[black, dashed, line width=0.8pt] ([xshift=4pt]l3.east) -- ++(0.3cm,0) coordinate (de);
    \node[right=1pt of de, font=\scriptsize] {\noprior{}};
}};
\end{tikzpicture}%
}
\caption{\small Debate consensus rate (CR), accuracy (ACC), and Consensus Correctness Rate (CCR) under \neutral{} ($s_{ij}=0$), by dataset, model group, and relation type. Values are percentages. Dashed lines show the corresponding \noprior{} baselines. Panel labels shorten OpinionQA, MMLU-Pro, and GPQA-Diamond to ``Opin.'', ``MMLU'', and ``GPQA'', respectively.}
\label{fig:neutral_baseline_debate}
\end{figure}

On objective QA, fully positive priors yield lower final majority-vote accuracy than \noprior{} in 10 of the 12 model-group--dataset--relation comparisons, match it in one, and exceed it in one (Fig.~\ref{fig:acc_comparison}). Accuracy improves from all-negative to all-positive relations in all six open-weight comparisons, whereas the GPT changes go in both directions. Positivity can therefore improve accuracy relative to negative framing while still underperforming the unmodified system.

The consensus results explain why agreement must be evaluated separately from accuracy. CR rises from all-negative to all-positive relations in all six open-weight objective-QA comparisons, but in only one of the six GPT comparisons (Fig.~\ref{fig:cr_comparison}). On MMLU-Pro, CCR is lower at the all-positive endpoint than at the all-negative endpoint in all six team--relation combinations; five of the six all-positive conditions also remain below \noprior{} (Fig.~\ref{fig:consensus_with_correction}). On GPQA-Diamond, CCR increases between the endpoints under \emph{Attitude} and \emph{Trust} for both teams, but decreases under \emph{Influence}. More frequent consensus can therefore coexist with less accurate consensus, depending on the team, dataset, and relation type.

The open-weight MMLU-Pro \emph{Influence} condition illustrates this directly. From $\lambda=0$ to $\lambda=1$, CR increases from 33.8\% to 76.9\%, a paired difference of $+43.1$ percentage points (95\% CI: $[+37.8,+48.2]$), while CCR decreases from 66.9\% to 58.4\%, a difference of $-8.5$ points (95\% CI: $[-15.8,-1.0]$). The team reaches consensus more often, but a smaller share of those consensuses is correct.

We examine how initial disagreement ends in the same 1,000-question MMLU-Pro experiment with the open-weight team under \emph{Influence} (Tab.~\ref{tab:answer-transitions}).

In this setting, more positive relations increase the fraction of initially disagreeing questions that end in unanimous agreement on a wrong answer. Changes in correcting a wrong initial majority or turning a correct initial majority into wrong consensus remain uncertain: both CIs span zero. Additional metric definitions and bootstrap details are provided in App.~\ref{app:answer-transitions}.
CR should therefore be reported together with accuracy and CCR on objective tasks.

The increase in consensus persists under both tested prompt changes. On the first 100 MMLU-Pro questions for the open-weight team, all-positive \emph{Influence} produces higher CR than all-negative \emph{Influence} with the original wording, a meaning-preserving paraphrase, and reordered relation statements. The increases are $+31.9$, $+39.6$, and $+43.3$ percentage points, respectively, and all three paired 95\% CIs exclude zero (App.~Tab.~\ref{tab:prompt-robustness}). The estimated size of the increase varies with wording and statement order. App.~\ref{app:prompt-robustness} provides the setup and prompt changes.


Relation type matters most clearly for consensus correctness. Among fully positive priors, \emph{Influence} yields the lowest CCR of the three relation types for both teams on both objective-QA datasets (Fig.~\ref{fig:consensus_with_correction}). Its direct emphasis on responding to others' recommendations may help explain this sensitivity. Across relation types, fully negative priors produce a larger proportional decline from \noprior{} in CR than in ACC in 10 of the 12 comparisons (App.~Fig.~\ref{fig:cr_acc_discussion}). Relational framing can therefore affect agreement and final-answer accuracy by substantially different amounts.

Under \neutral{}, both CR and ACC are below \noprior{} in all 12 objective-QA comparisons, while CCR changes in both directions (Fig.~\ref{fig:neutral_baseline_debate}). On OpinionQA, neutral \emph{Attitude} and \emph{Trust} raise CR for both teams, while neutral \emph{Influence} lowers it. Explicit neutrality therefore changes behavior in ways that depend on the task and relation type, and should be evaluated separately from omitting relational framing.

\section{Design Implications}
\label{sec:design_implication}



Choose the task's primary metric before selecting a relational prompt, and compare every candidate with \noprior{}. For objective QA that returns an answer to every question, select prompts using validation accuracy and report CR and CCR alongside it. These additional metrics show how often agents agree and how often their agreement is correct. Keep \noprior{} as the default for accuracy-centric tasks unless a relational prior improves the primary validation metric. For tasks that reward consensus, compare CR; for cooperative governance, compare sustainability outcomes. Our results show why the baseline matters: improving on negative relations and improving on the unmodified system are different outcomes.

Validate relation type and topology with the target model. In fully positive GovSim configurations, \emph{Attitude} has the smallest worst-case loss, while \emph{Influence} spans the largest gains and losses. Chain spans a wider range across models and relation types than star or tree. In objective debate, fully positive \emph{Influence} gives the lowest CCR of the three relation types for both teams on both datasets. These results favor evaluating the specific model--relation--topology combination rather than choosing a relational prompt from its intended social meaning alone.

Evaluate \neutral{} separately: stating neutrality changes outcomes relative to omitting relational framing. Use all-negative and all-positive networks as boundary conditions, and include mixed signs, relation-type variation, and applicable sparse topologies to test sensitivity to relational framing. These comparisons make relational priors useful both as task-specific interventions and as diagnostic probes of how social context changes group behavior.
\section{Conclusion}
\label{sec:conclusion}

We studied explicit relational priors in LLM-based multi-agent systems using a signed-network prompt intervention that keeps task protocols fixed. More positive relations generally improve sustainability within the GovSim relational-prior sweep and increase consensus on subjective questions. On objective QA, fully positive priors usually remain below no-prior accuracy, and some conditions produce more frequent but less accurate consensus. The central distinction is therefore between making agents coordinate or agree and improving the task's actual outcome. Model backbone, relation type, topology, and explicit neutrality all affect that distinction. Relational priors provide task-specific interventions and diagnostic probes of convergence pressure; their value should be established against a no-prior baseline using the task's primary metric. For accuracy-centric tasks, no-prior prompting remains the default unless validation supports a relational prior.

\section*{Limitations}
We study prompt-level relations that remain fixed, symmetric, and common knowledge throughout each interaction. This design isolates relational framing while holding the task protocol and communication graph fixed. Private, asymmetric, learned, or evolving relations may produce different effects; directed authority relations also require separate study.

Our experiments cover two task families: commons governance and three-agent debate. These settings contrast coordination, consensus, and correctness, and our findings describe the tested tasks, models, and relational structures. Larger teams, tool-using agents, persistent memory, human-agent interaction, and task-specific role hierarchies may change how relational priors affect behavior and task performance. Relational prompting therefore needs validation in each target setting.

\bibliography{custom}

\appendix
\begin{figure}[h]
\centering
\begin{tcolorbox}[
    colback=gray!5,
    colframe=gray!50,
    title=\textbf{Agent System Prompt in GovSim},
    fonttitle=\small,
    left=4pt,
    right=4pt,
    top=2pt,
    bottom=2pt,
    width=\columnwidth
]
\small\ttfamily
You are John. [Task-related prompt]

\medskip
You and other fishermen hold various attitudes towards each other. Attitude can be quantified using a score within the range of [-1, 1], where 1 represents the strongest positive attitude, such as kindness and affection, and -1 represents the most negative attitude, such as hostility and resentment.

\medskip
John's attitude score towards Kate is 1. Kate's attitude score towards John is also 1. John's attitude score towards Jack is -1. Jack's attitude score towards John is also -1. Kate's attitude score towards Jack is -1. Jack's attitude score towards Kate is also -1.

\medskip
You make all your decisions considering the attitudes among all of you in the group.
\end{tcolorbox}
\caption{\small Example GovSim system prompt for agent John under the \emph{Attitude} relation type. It combines the task instructions with the relation definition and verbalized signed edges of the relation network.}
\label{box:prompt_template}
\end{figure}

\section{GovSim Details}

\subsection{Simulation Dynamics}
\label{sec:govsim_details_dynamics}
\paragraph{Resource dynamics.}
At the start of month $t$, the shared stock $h(t)$ is available. Each simulation lasts $T=12$ months or until the stock collapses. A collapse occurs when $h(t)$ falls below a small constant $C=5$, at which point no further extraction is possible. The sustainability threshold $f(t)$ is the maximum combined harvest that leaves the next month's stock at least as large as the current stock. With GovSim's doubling-regeneration dynamics, $f(t)=h(t)/2$. After harvest, the resource regenerates according to
\begin{equation}
    h(t+1)=\min\left\{2\left[h(t)-\sum_{i=1}^{n} r_{i,t}\right],100\right\},
\end{equation}
where $r_{i,t}$ is the quantity harvested by agent $i$ at month $t$. The remaining stock doubles, up to the capacity of 100. To maintain the resource, the group must therefore harvest no more than half of the current stock in total.

\paragraph{Monthly phases.}
Each month comprises two phases. In the \emph{harvest phase}, agents privately choose how much of the resource to extract, subject only to the available stock. Decisions are submitted simultaneously; after execution, each agent's choice and the remaining stock $h(t)-\sum_i r_{i,t}$ are disclosed to all agents. In the \emph{discussion phase}, agents converse in natural language. They may share beliefs about sustainability, negotiate, appeal to moral principles, or adjust their strategies. Relational priors supply context for these decisions and discussions; agents still choose their own words and harvest quantities. After discussion, the new stock $h(t+1)$ is computed as above. The process repeats until $t=T$ or the stock collapses.

\input{figures/fig12_topology}

\input{figures/fig08_govsim_5agents}

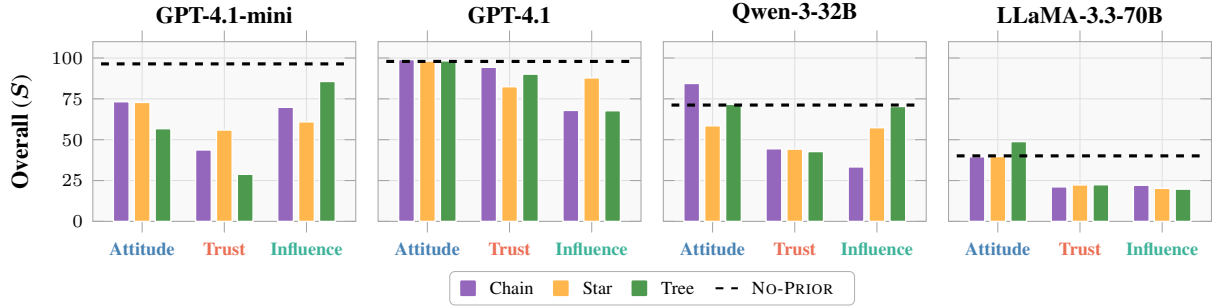
\begin{figure*}[t]
\centering
\definecolor{chaincolor}{RGB}{148,103,189}
\definecolor{starcolor}{RGB}{255,183,77}
\definecolor{treecolor}{RGB}{77,153,77}
\definecolor{attitudecolor}{RGB}{70,130,180}
\definecolor{trustcolor}{RGB}{235,110,85}
\definecolor{influencecolor}{RGB}{60,179,153}
\resizebox{\textwidth}{!}{%
\begin{tikzpicture}
\pgfplotsset{neutralfive/.style={
    width=0.32\textwidth, height=4cm, ybar, bar width=6pt,
    ymin=0, ymax=110, ytick={0,25,50,75,100},
    xtick={1,2,3}, xticklabels={{\textcolor{attitudecolor}{\textbf{Attitude}}},{\textcolor{trustcolor}{\textbf{Trust}}},{\textcolor{influencecolor}{\textbf{Influence}}}},
    xticklabel style={font=\scriptsize}, tick label style={font=\scriptsize},
    axis background/.style={fill=gray!5}, axis line style={gray!70}, grid=major,
    grid style={gray!25, line width=0.3pt}, every tick/.style={gray!70},
    enlarge x limits=0.3, clip=false, title style={font=\small, yshift=-1mm}
}}
\begin{axis}[neutralfive, name=plot1, title={\textbf{GPT-4.1-mini}}, ylabel={\textbf{Overall} ($\bm{S}$)}, ylabel style={font=\small}]
\addplot[fill=chaincolor, draw=white, line width=0.5pt] coordinates {(1,73.2) (2,43.7) (3,69.8)};
\addplot[fill=starcolor, draw=white, line width=0.5pt] coordinates {(1,72.8) (2,55.9) (3,60.9)};
\addplot[fill=treecolor, draw=white, line width=0.5pt] coordinates {(1,56.7) (2,28.8) (3,85.6)};
\draw[black, dashed, line width=1.2pt] (axis cs:0.5,96.4) -- (axis cs:3.5,96.4);
\end{axis}
\begin{axis}[neutralfive, name=plot2, at={($(plot1.east)+(0.3cm,0)$)}, anchor=west, title={\textbf{GPT-4.1}}, yticklabels={}]
\addplot[fill=chaincolor, draw=white, line width=0.5pt] coordinates {(1,99.0) (2,94.3) (3,67.9)};
\addplot[fill=starcolor, draw=white, line width=0.5pt] coordinates {(1,98.1) (2,82.4) (3,87.8)};
\addplot[fill=treecolor, draw=white, line width=0.5pt] coordinates {(1,98.2) (2,90.1) (3,67.7)};
\draw[black, dashed, line width=1.2pt] (axis cs:0.5,97.9) -- (axis cs:3.5,97.9);
\end{axis}
\begin{axis}[neutralfive, name=plot3, at={($(plot2.east)+(0.3cm,0)$)}, anchor=west, title={\textbf{Qwen-3-32B}}, yticklabels={}]
\addplot[fill=chaincolor, draw=white, line width=0.5pt] coordinates {(1,84.4) (2,44.4) (3,33.3)};
\addplot[fill=starcolor, draw=white, line width=0.5pt] coordinates {(1,58.5) (2,44.2) (3,57.3)};
\addplot[fill=treecolor, draw=white, line width=0.5pt] coordinates {(1,71.6) (2,42.7) (3,70.3)};
\draw[black, dashed, line width=1.2pt] (axis cs:0.5,71.2) -- (axis cs:3.5,71.2);
\end{axis}
\begin{axis}[neutralfive, name=plot4, at={($(plot3.east)+(0.3cm,0)$)}, anchor=west, title={\textbf{LLaMA-3.3-70B}}, yticklabels={}]
\addplot[fill=chaincolor, draw=white, line width=0.5pt] coordinates {(1,39.5) (2,21.1) (3,22.1)};
\addplot[fill=starcolor, draw=white, line width=0.5pt] coordinates {(1,39.7) (2,22.3) (3,20.2)};
\addplot[fill=treecolor, draw=white, line width=0.5pt] coordinates {(1,48.8) (2,22.4) (3,19.7)};
\draw[black, dashed, line width=1.2pt] (axis cs:0.5,40.1) -- (axis cs:3.5,40.1);
\end{axis}
\path (plot1.south west) -- (plot4.south east) coordinate[midway] (figcenter);
\node[draw=gray!60, fill=white, rounded corners=2pt, inner xsep=4pt, inner ysep=2pt, line width=0.4pt] at ([yshift=-0.9cm]figcenter |- plot1.south) {\tikz[baseline=-0.5ex]{
    \node[fill=chaincolor, rounded corners=0.5pt, minimum width=6pt, minimum height=6pt, inner sep=0pt] (c1) {};
    \node[right=1pt of c1, font=\scriptsize] (l1) {Chain};
    \node[fill=starcolor, rounded corners=0.5pt, minimum width=6pt, minimum height=6pt, inner sep=0pt, right=3pt of l1] (c2) {};
    \node[right=1pt of c2, font=\scriptsize] (l2) {Star};
    \node[fill=treecolor, rounded corners=0.5pt, minimum width=6pt, minimum height=6pt, inner sep=0pt, right=3pt of l2] (c3) {};
    \node[right=1pt of c3, font=\scriptsize] (l3) {Tree};
    \draw[black, dashed, line width=0.8pt] ([xshift=4pt]l3.east) -- ++(0.3cm,0) coordinate (de);
    \node[right=1pt of de, font=\scriptsize] {\noprior{}};
}};
\end{tikzpicture}%
}
\caption{\small Five-agent GovSim overall score $S$ under \neutral{}, by model backbone, relation type, and topology. Dashed lines show the corresponding \noprior{} baselines.}
\label{fig:neutral_baseline_5agents}
\end{figure*}

\subsection{Evaluation}
\label{sec:govsim_details_metrics}
We use the five metrics defined by the original GovSim benchmark.

\begin{itemize}[topsep=0pt, partopsep=0pt, itemsep=2pt, parsep=0pt, leftmargin=12pt]
    \item \textbf{Survival time} ($m$) --- The survival time is the longest number of months during which the shared resource remains above the collapse threshold $C$: $m=\max(\{t\in\mathbb{N}\mid h(t)>C\})$.

    \item \textbf{Survival rate} ($q$) --- Proportion of runs that achieve maximum survival time, i.e., $m=12$: $q=\frac{\#\{m=12\}}{\#\text{runs}}$.

    \item \textbf{Total gain} ($R$) --- Average total gain per agent over the survival period: $R=\frac{1}{|I|}\sum_{i=1}^{|I|}R_i$, where $R_i=\sum_{t=1}^{T}r_{i,t}$ is the total gain of agent $i$.

    \item \textbf{Inequality} ($e$) --- Gini-based equality score across the total gains of each agent: $e=1-\frac{\sum_{i=1}^{|I|}\sum_{j=1}^{|I|}|R_i-R_j|}{2|I|\sum_{i=1}^{|I|}R_i}$.

    \item \textbf{Over-usage} ($o$) --- Amount of unsustainable behavior across a simulation. It is the percentage of actions across the simulation that exceed the sustainability threshold: $o=\frac{\sum_{i=1}^{|I|}\sum_{t=1}^{T}\mathbf{1}(r_{i,t}>f(t))}{|I|\cdot m}$.
\end{itemize}

\noindent We also use the aggregate score $S$ to integrate the original five metrics. We first normalize each metric by its theoretical maximum value, making sure that larger values correspond to better performance. For example, for inequality/equality $e$, the normalized metric is $\tilde{e}=1-e$ when $e$ is represented as inequality. We then take a weighted linear mean:
\begin{equation}
    S=w_1\tilde{m}+w_2\tilde{q}+w_3\tilde{R}+w_4\tilde{e}+w_5\tilde{o},
\end{equation}
where $w_1=w_2=w_5=\frac{2}{8}$ and $w_3=w_4=\frac{1}{8}$, since survival time, survival rate, and over-usage directly assess system sustainability.

\begin{table*}[t]
\centering
\resizebox{0.65\linewidth}{!}{%
\small
\begin{tabular}{llcc}
\toprule
Model & Relation & $\Delta S$ & 95\% bootstrap CI \\
\midrule
GPT-4.1-mini & Attitude  & $+68.4$ & $[+43.2,+81.5]$ \\
GPT-4.1-mini & Trust     & $+39.0$ & $[+10.1,+67.7]$ \\
GPT-4.1-mini & Influence & $+78.1$ & $[+76.7,+81.1]$ \\
GPT-4.1      & Attitude  & $+81.4$ & $[+80.8,+82.2]$ \\
GPT-4.1      & Trust     & $+80.7$ & $[+79.5,+82.1]$ \\
GPT-4.1      & Influence & $+82.2$ & $[+81.6,+82.8]$ \\
Qwen-3-32B   & Attitude  & $+24.7$ & $[-13.7,+63.4]$ \\
Qwen-3-32B   & Trust     & $+50.2$ & $[+21.8,+78.4]$ \\
Qwen-3-32B   & Influence & $-21.1$ & $[-52.3,+15.8]$ \\
LLaMA-3.3-70B-Instruct & Attitude  & $+36.8$ & $[+7.7,+65.7]$ \\
LLaMA-3.3-70B-Instruct & Trust     & $+36.5$ & $[+7.8,+65.4]$ \\
LLaMA-3.3-70B-Instruct & Influence & $+38.1$ & $[+8.8,+67.1]$ \\
\bottomrule
\end{tabular}
}
\caption{\small Three-agent GovSim all-positive minus all-negative score differences, $\Delta S=S_{\lambda=1}-S_{\lambda=0}$, with 95\% complete-run bootstrap confidence intervals. Each endpoint contains five runs. Differences are measured in score points.}
\label{tab:govsim-bootstrap}
\end{table*}

\subsection{GovSim Bootstrap Intervals}
\label{app:govsim-uncertainty}

For each model and relation type in the complete three-agent setting, we estimate the all-positive minus all-negative score difference, $\Delta S = S_{\lambda=1}-S_{\lambda=0}$. Each condition contains five simulation runs. We construct a bootstrap sample by selecting five runs with replacement from each condition, independently, keeping all agents and months within each selected run together. We then recompute $S$ for each condition and calculate their difference. With five runs per condition, we can enumerate all possible ordered resamples and calculate the score difference for every pair of endpoint resamples. The 2.5th and 97.5th percentiles of these differences give the 95\% bootstrap confidence interval. Ten of the twelve intervals lie entirely above zero, supporting higher sustainability scores under all-positive relations in those comparisons. For Qwen-3-32B under \emph{Attitude} and \emph{Influence}, the intervals span zero and the direction of the difference remains uncertain. Tab.~\ref{tab:govsim-bootstrap} reports all differences and intervals.


\section{Debate Details}

\subsection{Debate Dynamics}
\label{sec:debate_formalization}
We follow the AutoGen debate protocol.\footnote{\url{https://microsoft.github.io/autogen/stable/user-guide/core-user-guide/design-patterns/multi-agent-debate.html}} We model a $K$-agent, $T$-round debate for question $x_n$. Let $z_{i,n}^{(t)}$ denote the complete textual response produced by agent $i\in\{1,\ldots,K\}$ at round $t$, including its selected answer and explanation. Let $\mathcal{Y}_n$ be the answer-option set for $x_n$, and let
\begin{equation}
    a_{i,n}^{(t)}=\operatorname{Extract}\!\left(z_{i,n}^{(t)}\right)\in\mathcal{Y}_n
\end{equation}
denote the extracted answer option. We write $Z_n^{(t)}=(z_{1,n}^{(t)},\ldots,z_{K,n}^{(t)})$ for the agents' responses at round $t$, and $Z_{-i,n}^{(t)}=(z_{j,n}^{(t)})_{j\neq i}$ for the responses of all agents other than $i$.

In round 1, agents answer independently given the question and their relational prompts:
\begin{equation}
    z_{i,n}^{(1)}
    =\mathrm{LLM}_i\!\left(x_n\mid\Pi_i(r,G_R)\right).
\end{equation}
For each revision round $t=2,\ldots,T$, agent $i$ retains its accumulated dialogue history $H_{i,n}^{(t-1)}$ and is additionally shown the other agents' responses from the preceding round:
\begin{equation}
    z_{i,n}^{(t)}
    =\mathrm{LLM}_i\!\left(
        x_n\mid\Pi_i(r,G_R),H_{i,n}^{(t-1)},Z_{-i,n}^{(t-1)}
    \right).
\end{equation}
The debate trajectory for question $x_n$ is the sequence $\{Z_n^{(t)}\}_{t=1}^{T}$. We use greedy decoding for all models and three answer rounds in total: one initial answer followed by two revisions.

\subsection{Evaluation}
\label{sec:debate_evaluation_details}
All debate metrics are computed from the extracted answer options $a_{i,n}^{(t)}$. For a dataset of $N$ questions, question $x_n$ is in full consensus at round $t$ if $a_{1,n}^{(t)}=\cdots=a_{K,n}^{(t)}$. Let
\begin{equation}
    I=\left\{n\in\{1,\ldots,N\}:
    \exists\,i,j\ \text{such that}\ a_{i,n}^{(1)}\neq a_{j,n}^{(1)}
    \right\}
\end{equation}
be the set of questions that begin with disagreement, and let
\begin{equation}
    C=\left\{n\in I:
    a_{1,n}^{(T)}=\cdots=a_{K,n}^{(T)}
    \right\}
\end{equation}
be the subset that reaches full consensus in the final round. We construct $I$ and $C$ separately for each condition because relational prompts can change both initial and final answers. The final-round consensus rate is the fraction of initially disagreeing questions that reach consensus:
\begin{equation}
    \mathrm{CR}=\frac{|C|}{|I|}.
\end{equation}

For subjective tasks, we use OpinionQA, a dataset derived from opinion polls and designed to evaluate alignment with human opinions on subjective issues. These questions have no universally correct answers, so CR measures how often initially disagreeing agents reach unanimous agreement.

For objective tasks, we use MMLU-Pro and GPQA-Diamond and apply the same definition of CR. We obtain the final debate answer by majority voting over the agents' final-round answer options:
\begin{equation}
    \hat{y}_n
    =\arg\max_{y\in\mathcal{Y}_n}
    \sum_{i=1}^{K}\mathbf{1}\!\left[a_{i,n}^{(T)}=y\right].
\end{equation}
Since each question has a ground-truth answer $y_n$, we report the accuracy of $\hat{y}_n$ over all $N$ questions. For each $n\in C$, let $c_n=a_{1,n}^{(T)}=\cdots=a_{K,n}^{(T)}$ denote the final consensus answer. The \emph{Consensus Correctness Rate} (CCR) is the fraction of these newly formed consensuses that are correct:
\begin{equation}
    \mathrm{CCR}
    =\frac{|C_{\mathrm{corr}}|}{|C|},
    \qquad
    C_{\mathrm{corr}}=\left\{n\in C:c_n=y_n\right\}.
\end{equation}
Thus, ACC measures final-answer accuracy on the full dataset, CR measures how often initial disagreement ends in consensus, and CCR measures the correctness of those consensuses.

\subsection{Debate Bootstrap Intervals}
\label{app:debate-uncertainty}

For the open-weight MMLU-Pro team under \emph{Influence}, we estimate the all-positive minus all-negative differences in CR and CCR using a paired question bootstrap. We sample 1,000 questions with replacement from the evaluated questions and use the same sampled questions in both conditions, retaining each question's recorded responses from all agents across all rounds. For each condition, we recompute CR as the fraction of initially disagreeing questions that reach final consensus, and CCR as the fraction of those consensuses that are correct. Initial answers and final consensus can differ between conditions, so the questions contributing to each rate are determined separately. We then subtract the all-negative rate from the all-positive rate. We repeat this procedure 100,000 times using random seed 0 and take the 2.5th and 97.5th percentiles of the resulting differences as the endpoints of the 95\% bootstrap confidence interval. The differences are $+43.1$ percentage points for CR (95\% CI: $[+37.8,+48.2]$) and $-8.5$ points for CCR (95\% CI: $[-15.8,-1.0]$). In this setting, all-positive relations produce consensus more often, but a smaller fraction of those consensuses is correct.


\subsection{Initial-to-Final Answer Transitions}
\label{app:answer-transitions}

We complement CCR with transitions that describe which initial answer patterns end in correct or wrong consensus. We analyze the open-weight MMLU-Pro team under \emph{Influence}, comparing the all-negative ($\lambda=0$) and all-positive ($\lambda=1$) conditions used in Sec.~\ref{sec:debate_results}. We use the paired question-level bootstrap procedure described in App.~\ref{app:debate-uncertainty}.

We report three question-level measures. \emph{Wrong-majority correction} is the fraction of initial 2--1 splits with a wrong majority that end in 3--0 correct consensus. \emph{Correct-majority failure} is the fraction of initial 2--1 splits with a correct majority that end in 3--0 wrong consensus. \emph{Wrong Consensus Formation} (WCF) is the fraction of all initially non-unanimous questions, including 1--1--1 splits, that end in 3--0 wrong consensus. Each rate uses the questions with its specified initial answer pattern, determined separately in each condition. The comparison is therefore between all-positive and all-negative relations within each measure.
For each measure, we estimate the all-positive minus all-negative difference
using 100,000 paired question-level bootstrap samples. The same
question indices are used at both endpoints, and the corresponding
conditional denominator is recomputed in every sample.

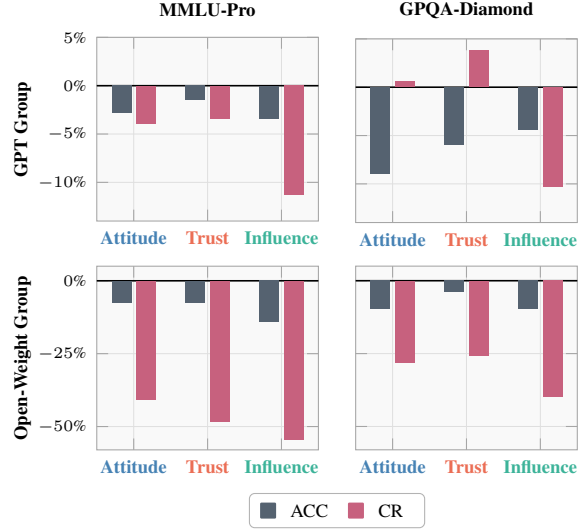
\begin{figure}[t]
\centering
\resizebox{1\columnwidth}{!}{%
\begin{tikzpicture}
\definecolor{attitudecolor}{RGB}{70,130,180}
\definecolor{trustcolor}{RGB}{235,110,85}
\definecolor{influencecolor}{RGB}{60,179,153}
\definecolor{acccolor}{RGB}{85,98,112}
\definecolor{crcolor}{RGB}{199,97,126}

\begin{axis}[
    name=plot11,
    width=4.5cm,
    height=4cm,
    title={\textbf{MMLU-Pro}},
    title style={font=\scriptsize, yshift=-2pt},
    ylabel={\textbf{GPT Group}},
    every axis y label/.style={at={(axis description cs:-0.25,0.5)}, rotate=90, anchor=south},
    ylabel style={font=\scriptsize},
    ymin=-14, ymax=5,
    xmin=0.5, xmax=3.5,
    ytick={-10,-5,0,5},
    yticklabel={\pgfmathprintnumber{\tick}\%},
    xtick={1,2,3},
    xticklabels={{\textbf{\textcolor{attitudecolor}{Attitude}}},{\textbf{\textcolor{trustcolor}{Trust}}},{\textbf{\textcolor{influencecolor}{Influence}}}},
    xticklabel style={font=\scriptsize},
    tick label style={font=\tiny},
    axis background/.style={fill=gray!5},
    axis line style={gray!70},
    grid=major,
    grid style={gray!25, line width=0.3pt},
    every tick/.style={gray!70},
    clip=false,
]
\draw[black, line width=0.6pt] (axis cs:0.5,0) -- (axis cs:3.5,0);
\fill[acccolor] (axis cs:0.70,0) rectangle (axis cs:0.97,-2.8);
\fill[crcolor] (axis cs:1.03,0) rectangle (axis cs:1.30,-3.9);
\fill[acccolor] (axis cs:1.70,0) rectangle (axis cs:1.97,-1.5);
\fill[crcolor] (axis cs:2.03,0) rectangle (axis cs:2.30,-3.4);
\fill[acccolor] (axis cs:2.70,0) rectangle (axis cs:2.97,-3.4);
\fill[crcolor] (axis cs:3.03,0) rectangle (axis cs:3.30,-11.3);
\end{axis}

\begin{axis}[
    name=plot12,
    at={(plot11.outer east)},
    xshift=0.1cm,
    anchor=outer west,
    width=4.5cm,
    height=4cm,
    title={\textbf{GPQA-Diamond}},
    title style={font=\scriptsize, yshift=-2pt},
    ymin=-14, ymax=5,
    xmin=0.5, xmax=3.5,
    ytick={-10,-5,0,5},
    yticklabels={},
    xtick={1,2,3},
    xticklabels={{\textbf{\textcolor{attitudecolor}{Attitude}}},{\textbf{\textcolor{trustcolor}{Trust}}},{\textbf{\textcolor{influencecolor}{Influence}}}},
    xticklabel style={font=\scriptsize},
    tick label style={font=\tiny},
    axis background/.style={fill=gray!5},
    axis line style={gray!70},
    grid=major,
    grid style={gray!25, line width=0.3pt},
    every tick/.style={gray!70},
    clip=false,
]
\draw[black, line width=0.6pt] (axis cs:0.5,0) -- (axis cs:3.5,0);
\fill[acccolor] (axis cs:0.70,0) rectangle (axis cs:0.97,-8.9);
\fill[crcolor] (axis cs:1.03,0) rectangle (axis cs:1.30,0.6);
\fill[acccolor] (axis cs:1.70,0) rectangle (axis cs:1.97,-5.9);
\fill[crcolor] (axis cs:2.03,0) rectangle (axis cs:2.30,3.8);
\fill[acccolor] (axis cs:2.70,0) rectangle (axis cs:2.97,-4.4);
\fill[crcolor] (axis cs:3.03,0) rectangle (axis cs:3.30,-10.3);
\end{axis}

\begin{axis}[
    name=plot21,
    at={(plot11.outer south)},
    yshift=-0.15cm,
    anchor=outer north,
    width=4.5cm,
    height=4cm,
    ylabel={\textbf{Open-Weight Group}},
    every axis y label/.style={at={(axis description cs:-0.25,0.5)}, rotate=90, anchor=south},
    ylabel style={font=\scriptsize},
    ymin=-58, ymax=5,
    xmin=0.5, xmax=3.5,
    ytick={-50,-25,0},
    yticklabel={\pgfmathprintnumber{\tick}\%},
    xtick={1,2,3},
    xticklabels={{\textbf{\textcolor{attitudecolor}{Attitude}}},{\textbf{\textcolor{trustcolor}{Trust}}},{\textbf{\textcolor{influencecolor}{Influence}}}},
    xticklabel style={font=\scriptsize},
    tick label style={font=\tiny},
    axis background/.style={fill=gray!5},
    axis line style={gray!70},
    grid=major,
    grid style={gray!25, line width=0.3pt},
    every tick/.style={gray!70},
    clip=false,
]
\draw[black, line width=0.6pt] (axis cs:0.5,0) -- (axis cs:3.5,0);
\fill[acccolor] (axis cs:0.70,0) rectangle (axis cs:0.97,-7.5);
\fill[crcolor] (axis cs:1.03,0) rectangle (axis cs:1.30,-40.8);
\fill[acccolor] (axis cs:1.70,0) rectangle (axis cs:1.97,-7.5);
\fill[crcolor] (axis cs:2.03,0) rectangle (axis cs:2.30,-48.2);
\fill[acccolor] (axis cs:2.70,0) rectangle (axis cs:2.97,-14.1);
\fill[crcolor] (axis cs:3.03,0) rectangle (axis cs:3.30,-54.4);
\end{axis}

\begin{axis}[
    name=plot22,
    at={(plot21.outer east)},
    xshift=0.1cm,
    anchor=outer west,
    width=4.5cm,
    height=4cm,
    ymin=-58, ymax=5,
    xmin=0.5, xmax=3.5,
    ytick={-50,-25,0},
    yticklabels={},
    xtick={1,2,3},
    xticklabels={{\textbf{\textcolor{attitudecolor}{Attitude}}},{\textbf{\textcolor{trustcolor}{Trust}}},{\textbf{\textcolor{influencecolor}{Influence}}}},
    xticklabel style={font=\scriptsize},
    tick label style={font=\tiny},
    axis background/.style={fill=gray!5},
    axis line style={gray!70},
    grid=major,
    grid style={gray!25, line width=0.3pt},
    every tick/.style={gray!70},
    clip=false,
]
\draw[black, line width=0.6pt] (axis cs:0.5,0) -- (axis cs:3.5,0);
\fill[acccolor] (axis cs:0.70,0) rectangle (axis cs:0.97,-9.8);
\fill[crcolor] (axis cs:1.03,0) rectangle (axis cs:1.30,-28.1);
\fill[acccolor] (axis cs:1.70,0) rectangle (axis cs:1.97,-3.9);
\fill[crcolor] (axis cs:2.03,0) rectangle (axis cs:2.30,-25.6);
\fill[acccolor] (axis cs:2.70,0) rectangle (axis cs:2.97,-9.8);
\fill[crcolor] (axis cs:3.03,0) rectangle (axis cs:3.30,-39.8);
\end{axis}

\path (plot21.south) -- (plot22.south) coordinate[midway] (midbot);
\node[draw=gray!60, fill=white, rounded corners=2pt, inner xsep=4pt, inner ysep=2pt, line width=0.4pt] at ([yshift=-0.8cm]midbot) {\tikz[baseline=-0.5ex]{
    \node[fill=acccolor, rounded corners=0.5pt, minimum width=6pt, minimum height=6pt, inner sep=0pt] (c1) {};
    \node[right=1pt of c1, font=\scriptsize] (l1) {ACC};
    \node[fill=crcolor, rounded corners=0.5pt, minimum width=6pt, minimum height=6pt, inner sep=0pt, right=3pt of l1] (c2) {};
    \node[right=1pt of c2, font=\scriptsize] (l2) {CR};
}};
\end{tikzpicture}%
}
\caption{\small Percentage changes in debate consensus rate (CR) and accuracy (ACC) at $\lambda=0$ relative to \noprior{}, by model group, dataset, and relation type. Solid lines mark zero change.}
\label{fig:cr_acc_discussion}
\end{figure}

\begin{table*}[t]
\centering
\resizebox{0.65\linewidth}{!}{%
\small
\begin{tabular}{lcccc}
\toprule
Prompt variant
& all $-1$
& all $+1$
& $\Delta$CR (points)
& Paired 95\% CI \\
\midrule
Original
& 42.0\%
& 73.9\%
& $+31.9$
& $[+13.9,+49.3]$ \\
Paraphrased
& 30.8\%
& 70.4\%
& $+39.6$
& $[+22.8,+55.9]$ \\
Reordered
& 33.3\%
& 76.6\%
& $+43.3$
& $[+25.9,+59.7]$ \\
\bottomrule
\end{tabular}
}
\caption{\small Prompt wording and statement-order check for the open-weight team under \emph{Influence} on the 100-question MMLU-Pro subset. CR is measured among questions with initial disagreement. $\Delta$CR is all-positive minus all-negative, in percentage points. All three formulations produce higher CR under all-positive relations.}
\label{tab:prompt-robustness}
\end{table*}

Tab.~\ref{tab:answer-transitions} in Sec.~\ref{sec:debate_results} reports the transition rates, endpoint differences, and confidence intervals.



\section{Consensus and Accuracy under Fully Negative Priors}
\label{sec:more_debate_analysis}
Relative to \noprior{}, fully negative priors produce a larger proportional decline in consensus than in accuracy in 10 of the 12 comparisons (Fig.~\ref{fig:cr_acc_discussion}). We measure relative percentage change as $100(M_{\lambda=0}-M_{\mathrm{\noprior{}}})/M_{\mathrm{\noprior{}}}$ for $M\in\{\mathrm{CR},\mathrm{ACC}\}$. This pattern holds in all six open-weight conditions and all three GPT MMLU-Pro conditions, as well as GPT GPQA-Diamond under \emph{Influence}. The exceptions are GPT GPQA-Diamond under \emph{Attitude} and \emph{Trust}, where CR is slightly above \noprior{} while ACC is below it. For open-weight MMLU-Pro, CR decreases by 40.8--54.4\%, compared with 7.5--14.1\% for ACC. These are proportional changes from each metric's own baseline; the main debate figures instead report percentage-point differences.

\section{Prompt Robustness to Wording and Statement Order}
\label{app:prompt-robustness}

We test whether all-positive \emph{Influence} still produces more consensus than all-negative \emph{Influence} when the relational wording is paraphrased or the relation statements are reordered.

\subsection{Setup and Prompt Variants}
We used the first 100 questions from the main experiment's MMLU-Pro sample. The three-agent team comprised Mixtral-8x22B-Instruct-v0.1, Qwen-3-32B, and LLaMA-3.3-70B-Instruct. The debate protocol and decoding settings matched the main experiments. We evaluated six conditions: the original, paraphrased, and reordered \emph{Influence} prompts, each with all three mutual relations set to either $-1$ or $+1$.

The original condition used the main experiment's \emph{Influence} wording and listed the three mutual-pair blocks in the order A--B, A--C, B--C. The paraphrased condition replaced terms such as ``influence score,'' ``follow their lead,'' and ``contrarian influence'' with a response-score description in which an agent moves toward or away from another agent's recommendations. Agent identities, pair mappings, statement directions, and numerical signs were unchanged. The reordered condition retained the original text verbatim and changed only the pair-block order to B--C, A--C, A--B.

The relation-definition portion contained 82 words in both the original and paraphrased versions, and the six agent-specific relation statements contained 45 words in both. The variants are matched in word count; model-specific token counts may differ. The question prompts and subsequent-round decision prompts were unchanged across all conditions.

\subsection{Evaluation}
We report consensus rate (CR) for questions whose three initial answers were non-unanimous. A question counts as a final consensus only when all three final answers are the same option. For each prompt variant, we subtract the all-negative CR from the all-positive CR. We estimate 95\% bootstrap percentile intervals using 100,000 paired question-level bootstrap samples. The same resampled question indices are used across all six conditions, and the conditional denominators are recomputed in every sample.

\subsection{Results}
All-positive relations produce more consensus than all-negative relations with the original, paraphrased, and reordered prompts, and all three paired intervals exclude zero (Tab.~\ref{tab:prompt-robustness}). On this 100-question MMLU-Pro subset with the open-weight team, the increase in consensus therefore persists under both tested changes. The size of the increase varies with wording and statement order.

\section{Use of AI Assistants}
The manuscript uses AI assistance primarily for language editing and restructuring under the author's supervision. AI assistance was not used to originate the research idea, generate the experimental data, create the reported numerical results, or perform the evaluation.

\end{document}